\documentclass[11pt]{article}

\usepackage{amsmath,amsfonts,bm}

\def\eqref#1{equation~\ref{#1}}

\def\1{\bm{1}}

\DeclareMathAlphabet{\mathsfit}{\encodingdefault}{\sfdefault}{m}{sl}
\SetMathAlphabet{\mathsfit}{bold}{\encodingdefault}{\sfdefault}{bx}{n}

\usepackage{natbib}

\usepackage{hyperref}
\usepackage{url}
\usepackage{subcaption}
\usepackage{graphicx}
\usepackage{booktabs}
\usepackage{wrapfig}
\usepackage{etoc}
\usepackage{titletoc}
\usepackage[utf8]{inputenc}
\usepackage{textcomp}
\usepackage[most]{tcolorbox}
\usepackage{caption}
\usepackage{longtable}

\newtcolorbox{promptbox}[1][]{
  colback=cyan!3,
  colframe=cyan!80,
  fonttitle=\bfseries\small,
  title={#1},
  breakable,
  left=6pt, right=6pt, top=4pt, bottom=4pt,
  fontupper=\small\ttfamily,
  sharp corners,
  boxrule=0.5pt,
}

\title{Too Categorical to be Human: Emotion Concepts in LLMs and Humans}

\date{}
\author{%
\begin{minipage}{0.95\textwidth}
\centering
Sree Bhattacharyya$^{1}$ \quad
Evgenii Kuriabov$^{2}$ \quad
Lucas Craig$^{3}$ \quad
Tharun Dilliraj$^{3}$ \quad
Reginald B. Adams,~Jr.$^{4}$ \quad
Jia Li$^{2}$ \quad
James~Z.~Wang$^{1}$
\\[0.7em]
{\small
$^{1}$Department of Informatics and Intelligent Systems \\
$^{2}$Department of Statistics \\
$^{3}$Department of Computer Science and Engineering \\
$^{4}$Department of Psychology \\[0.2em]
The Pennsylvania State University, University Park, PA 16802, USA \\
\texttt{sreeb@psu.edu}
}
\end{minipage}%
}

\begin{document}

\maketitle

\begin{abstract}

Understanding human emotions is central to user-facing AI applications, safety alignment, and the simulation of human behavior. With evidence of emotional stimuli shaping high-stakes behavior in Large Language Models (LLMs), there is increasing interest in how models represent emotion concepts internally. Mechanistic accounts of these representations, however, cannot be compared directly against humans: emotion processing in humans is highly distributed and yields no equivalent neural representation. To understand whether LLMs internalize emotion concepts in a way similar to humans, we propose characterizing the abstract concept of an emotion using external behavioral signatures, which we term \textit{behavioral representations}. Using the theory of cognitive appraisals, which enables representing emotional situations along interpretable evaluative dimensions (e.g., how pleasant or effortful a situation is), we create a benchmark dataset of emotional scenarios spanning 15 emotion categories. We elicit behavioral representations of emotion concepts from LLMs and humans using our benchmark, and study their structural similarity. We find that LLMs represent emotion concepts more categorically, homogeneously, and determinately than humans, representing a single emotion concept with less internal diversity, and place different emotions further apart. Models from different families converge with one another on how they represent a given emotion significantly more than groups of humans do. The categorical structure of representations in LLMs is further robust to contextual variation, including with different task framing and demographic personas. Analyzing model checkpoints across different training stages, we also find that the discretized nature of representations appears after the mid-training stage itself and is unaffected by different post-training strategies. Through our results, we highlight a key difference in how LLMs behaviorally represent emotion concepts, curbing the subjectivity inherent to the human experience of emotions. More broadly, we argue that evaluating LLMs in subjective domains such as emotions requires moving beyond static prediction to assess the structure of what models represent.

\end{abstract}

\section{Introduction}
\label{sec:introduction}







Emotions are central to human experience, shaping how people perceive their circumstances, understand themselves, and act within the world. Equipping artificial systems with the ability to understand human emotions has been approached through several parallel directions: investigating how accurately a model \textit{predicts} the emotion or underlying reasoning associated with some emotionally-charged stimuli~\citep{wang2023unlocking, bhattacharyya2025evaluating, tak2023gpt, gandhi2024human}, exploring how emotional cues can influence general behavior in foundation models~\citep{mozikov2024eai, li2023large}, or the simulation of human-like emotional behavior~\citep{huang2024apathetic, bhattacharyya2026expressing}. A separate body of research asks how emotions are represented internally in models, particularly in Large Language Models (LLMs)~\citep{sofroniew2026emotion}.

\begin{figure}[t!]
    \centering
    \includegraphics[width=0.93\linewidth]{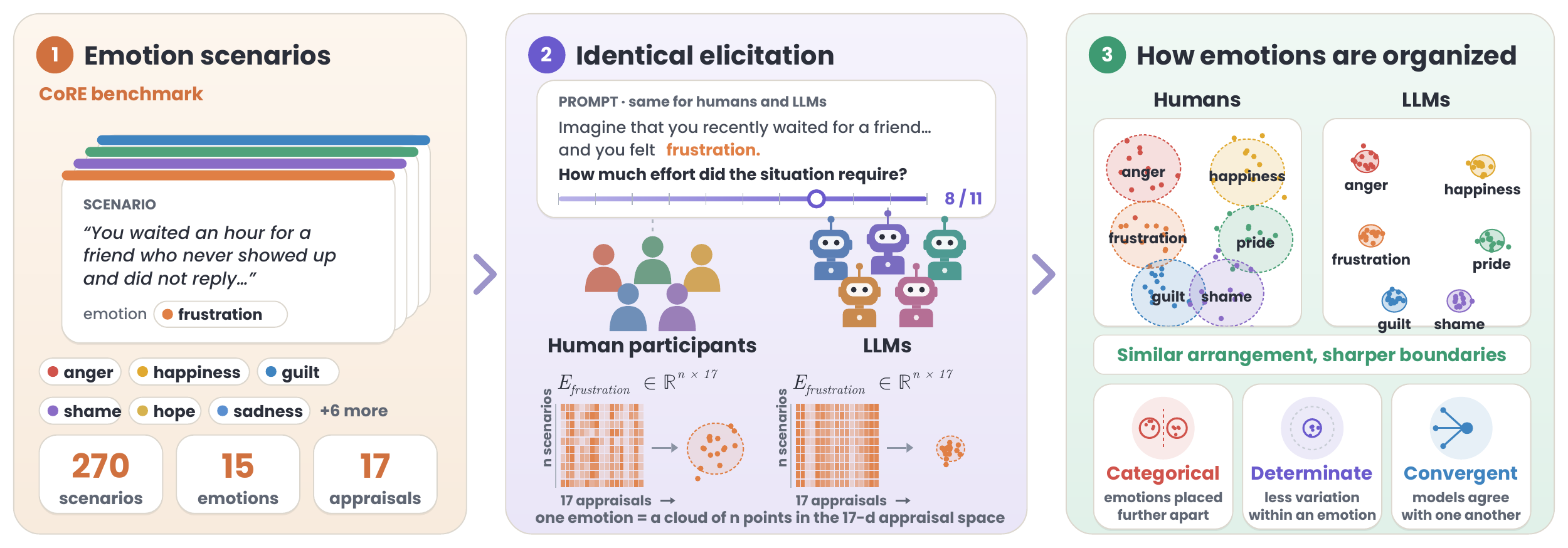}
    \caption{\textbf{Overview.} (1) CoRE contains 270 scenarios spanning 15 emotion categories, each rated on 17 appraisal dimensions. (2) Human participants and LLMs answer identical first-person appraisal prompts. For each agent, the scenarios of one emotion form a matrix $E_e \in \mathbb{R}^{n_e \times 17}$, that is, a cloud of $n_e$ points in appraisal space. (3) Comparing these clouds across emotions shows that LLMs arrange emotions similarly to humans but with sharper boundaries: emotion categories are further apart (categorical), vary less within a category (determinate), and are more similar across models (convergent). Clouds are schematic and not computed with the actual data.}
    \label{fig:overview}
\end{figure}

Studying emotion representations in LLMs is increasingly of interest, helping interpret whether models process emotions in a human-aligned manner. It is also consequential beyond emotion-related behavior, as emotion representations are shown to causally modulate sycophancy and reward hacking~\citep{sofroniew2026emotion}, emotional cues in prompts are found to influence model decisions in game-theoretic settings~\citep{mozikov2024eai} and improve performance in general tasks~\citep{li2023large}. Existing research on emotion representations has primarily focused on localizing \textit{where} in LLMs emotional information is encoded~\citep{tak2025mechanistic, wang2026do, reichman2026emotions}. 
However, localization of emotion representations does not make them interpretable in a human-like manner. Recent studies have shown that models arrange emotion concepts along broad affective axes such as valence and arousal, and form hierarchies resembling those posited in theories from human psychology~\citep{reichman2026emotions, okawa2026emergence}. Although such findings establish coarse theoretical plausibility, they do not establish finer structure, such as \textit{how} a single emotion concept is represented. More importantly, these findings lack a comparable characterization of whether model representations are aligned with humans responding to the same stimuli. 

Establishing finer human-model correspondence by mechanistically examining internal model activations is inherently challenging: emotions are produced as a distributed process in humans, with no singular, localized site to retrieve neural representations from~\citep{lindquist2012brain, lee2024more}. Further, emotional experience in humans is characterized by individual and contextual variability~\citep{siegel2018emotion, satpute2019default}, making the recovery of distinct emotion-specific neural signatures challenging. 

In this paper, to study finer human alignment of how LLMs understand emotions, we turn to \textit{behavioral representations}, where an abstract concept is characterized through external behavioral signatures~\citep{xu2025revealing, xu2024tip, du2025human}. Specifically, we use self-reports of cognitive appraisals of an emotion-related scenario to characterize how an LLM or humans (collectively referred to as agents henceforth) understand and internalize a given emotion concept. Cognitive appraisal theory~\citep{smith1985patterns, scherer1982emotion, scherer1997role} states that emotions arise from distinct ways of evaluating a situation, along interpretable axes (such as amount of effort required, amount of agency involved, etc.), providing a principled framework for representing distinct emotions along these axes. It is ideal for eliciting functionally comparable self-reports from both LLMs and humans, on the same set of emotion-specific stimuli. LLMs have also been shown to possess a functional, human-like understanding of common appraisal dimensions~\citep{tak2024gpt, gandhi2024human}, making their self-reports of cognitive appraisal comparable, interpretable, and reliable for characterizing individual emotion concepts. Throughout the work, we refer to appraisal-based characterizations of emotion concepts as \textit{behavioral representations}, interpreting them \textit{functionally}, and not as markers of neural or activation-level representations. 

Obtaining behavioral representations of emotions in a comparable format from both LLMs and humans, we ask: \textit{do LLMs and humans behaviorally represent individual emotion concepts in the same way?} In contrast to the more common behavioral evaluation target, which asks \textit{whether} LLMs' and humans' \textit{outputs} about emotions and appraisals are similar~\citep{tak2024gpt, Yongsatianchot2023investigating, gandhi2024human}, which we term \textit{directional alignment}, we study \textit{how} differently are emotion \textit{concepts} organized---evaluating \textit{structural} alignment. In particular, we focus on whether models capture contextual subjectivity inherent to an emotion concept~\citep{goldenberg2026large}, beyond differences due to individually-specific appraisals, whether different emotion concepts are organized similarly to humans, and how the structure of representations changes with varying amounts of context. 

Grounded in a seminal study from human psychology~\citep{smith1985patterns}, we introduce \textbf{CoRE} (\textbf{Co}gnitive \textbf{R}epresentations of \textbf{E}motions), a benchmark of emotion scenarios spanning 15 emotion categories, each characterized by 17 appraisal dimensions, and elicit first-person appraisals from human participants and eight LLMs under identical instructions (Fig.~\ref{fig:overview} provides an overview). Quantifying the structure of appraisal-based representations through a series of comprehensive, complementary analyses, we find that LLMs capture the broad appraisal geometry of human emotion concepts, but organize those concepts into more compressed, discrete, and homogeneous structures than humans. Specifically, our main findings are as follows: (1) \textbf{Discreteness of Emotion Concepts}: Despite shared coarse geometry, within an agent's own space, models place distinct instances of a single emotion significantly closer together, and distinct emotions significantly further apart, than humans do. (2) \textbf{Determinacy of Emotion Concepts}: Distinct emotion categories are distinguishable from an agent's own appraisals far more accurately for models than for humans. What appraisal axis a model \textit{explicitly} states distinguishes two emotions also closely matches the axis that indeed separates the same emotion concepts in its own representation space. (3) \textbf{Convergence of Emotion Concepts across Models}: Different language models agree with one another on the typical behavioral representation of a given emotion significantly more closely than different groups of humans. (4) \textbf{Invariance to Context and Emergence during Midtraining}: Limiting the emotional information presented, or inducing different kinds of demographic personas, changes the surface appraisals without aligning model representations to human-like
structure. Tracing model checkpoints for a single model family across training stages, we find that the categorical structure of representations is already present after midtraining rather than introduced by post-training.

Complementary to existing findings on homogeneity in LLMs at the \textit{output} level~\citep{jiang2025artificial, kirk2024understanding, ballestero2026strategic}, we highlight \textit{concept} level homogenization and discretization, for the domain of emotions, where variability and diversity are fundamental. Broadly, our work highlights the importance of evaluating \textit{structural} alignment beyond response-level alignment, and provides data and tools to progress the same.

\subsection{Related Works}

\textbf{Emotion representation in LLMs.} Most work on emotion in LLMs uses mechanistic interpretability, without reference to human data. Emotion concepts have been localized to layers~\citep{tak2025mechanistic}, to neurons and attention heads~\citep{wang2026do}, and to a low-dimensional subspace that generalizes across datasets and languages~\citep{reichman2026emotions}, with localization varying by emotion~\citep{shu2026syntax}. Intervening on these representations alters the expressed emotion but not the semantic content~\citep{wang2026do, reichman2026emotions}, reduces emotional hallucination in vision-language models~\citep{zhang2026interpreting}, and controls safety-relevant behaviors such as sycophancy and reward hacking~\citep{sofroniew2026emotion}. Reported links to human psychology, such as hierarchical emotion trees~\citep{okawa2026emergence} and subspaces matching theoretical dimensions~\citep{reichman2026emotions, sofroniew2026emotion}, concern broad trends rather than detailed correspondence between model and human organization. Because human emotion processing yields no internal representations comparable to model activations~\citep{lindquist2012brain}, we compare the two through explicit self-reports.

\textbf{Cognitive appraisals in LLMs.} Corpora of emotional narratives with appraisal annotations~\citep{zhan2023evaluating, yeo2023peace, troiano2022xenvent, sun2026carebench} enable testing how well models \textit{predict} appraisals from emotions and vice versa. LLMs predict appraisals nearly as well as human annotators~\citep{tak2024gpt, zhou2025modeling, gandhi2024human}, reproduce theoretically expected relations between appraisals and emotions~\citep{tak2025aware, thejll-madsen2026systematic}, and aid intention prediction~\citep{yeo2024beyond}, well-being analysis~\citep{liu2023psyam}, and cognitive reappraisal that experts judge acceptable~\citep{zhan2024large}. Their accuracy, however, varies by appraisal dimension~\citep{Yongsatianchot2023investigating} and is higher for third-party attributions than for first-person self-reports~\citep{tak2024gpt}. Intervening on appraisal representations also changes the emotion a model infers~\citep{tak2025mechanistic}. Whereas this work measures \textit{predictive accuracy} against a human reference, we use appraisals, an established characterization of emotion, to compare the \textit{structural properties} of emotion concepts in LLMs and humans.


\section{The Cognitive Representation of Emotions (CoRE) Benchmark}
\label{sec:benchmark}


We ground our benchmark in the theory of cognitive appraisal from human psychology~\citep{scherer1982emotion, lazarus1984stress, smith1985patterns}, which enables characterizing an emotion-laden scenario through a set of interpretable dimensions of evaluating the situation. Following a seminal study of emotional appraisal in humans~\citep{smith1985patterns}, our benchmark consists of scenarios spanning \textbf{15 distinct emotion categories} (Table~\ref{tab:app_emotions}), each characterized along \textbf{17 appraisal dimensions} (Table~\ref{tab:app_appraisals}). Although several human appraisal datasets exist, we curate a new benchmark for two main reasons (summary of comparison in Table~\ref{tab:appraisal_data_comparison}). First, most existing datasets record third-person appraisal: raters estimate the appraisals of \textit{someone else} experiencing an emotional scenario. These ratings capture how a rater infers another person's emotional response, instead of how they themselves appraise situations related to a given emotion. Second, in the few datasets that do record self-appraisal, the rater usually supplies the scenario as well, recalling it from memory. This creates several obstacles to a fair comparison with LLMs, such as: scenarios being skewed toward highly charged events~\citep{tak2023gpt}, not corresponding to a single emotion category that is self-reported~\citep{taylor2021examining}, or carrying context- or demographic-specific references~\citep{yeo2023peace}. Most critically, when each rater reports their own memory, no two raters can appraise the same scenario. A human-LLM comparison of the structure of emotions on such data would be unfair, as a single LLM would rate \textit{distinct} scenarios, which are each rated alone by a separate individual. Such a comparison would also make \textit{inherent, context-dependent} variability of a \textit{single emotion} concept indistinguishable from \textit{variability across different individuals}. We therefore construct a new appraisal-based dataset of emotion scenarios that are framed hypothetically rather than recalled, free of individual and demographic-specific references, with several human participants providing self-appraisals for the same scenario. We adopt a rigorous 4-stage pipeline for data curation, including human validation at multiple stages (Fig.~\ref{fig:app_core_curation}).

\begin{wraptable}{r}{0.5\textwidth}
\centering
\footnotesize
\setlength{\tabcolsep}{3.5pt}
\resizebox{\linewidth}{!}{
\begin{tabular}{@{}p{3.5cm}cp{2cm}@{}}
\toprule
\textbf{Dataset} & \textbf{Perspective} & \textbf{Ratings} \\
\midrule
x-enVent \citep{troiano2022xenvent}         & 3rd  & --    \\
crowd-enVent \citep{troiano2023dimensional} & dual & 1     \\
Covid-ET \citep{zhan2023evaluating}         & 3rd  & --    \\
PEACE-Reviews \citep{yeo2023peace}          & 3rd  & --    \\
\citet{yeo2025beyond}                             & 3rd  & --    \\
FGE \citep{skerry2015neural}                & 3rd  & --    \\
\citet{tak2025aware}                              & 1st  & 1     \\
\midrule
\textbf{CoRE} (ours) & \textbf{1st} & \textbf{$\approx$4} \\
\bottomrule
\end{tabular}}
\caption{Existing appraisal datasets compared with CoRE. \emph{Perspective} is the view the appraisal is made from (appraising self, or others). \emph{Ratings} is the number of independent self-appraisals collected per scenario.}
\label{tab:appraisal_data_comparison}
\end{wraptable}

In Stage 1, we use example scenarios self-reported by participants in~\citet{smith1985patterns} to create $\approx5$ scenarios for each of the 15 emotion categories. In Stage 2, we use GPT-4o~\citep {hurst2024gpt} to expand the same set to around $>20$ scenarios per emotion category. 
In Stage 3, we conduct a rigorous multi-rater verification (details in Appendix~\ref{sec:app_multi_rater_verification}). We recruit 8 annotators to validate that each chosen scenario is general, easily understood as a common daily-life situation, and sufficiently associated with the corresponding emotion category. Two independent raters verify each scenario, providing human-authored alternate scenarios wherever they reject a model-generated one. At the end of the multi-rater verification process, we retain 294 scenarios, each of which is further manually validated, edited, rephrased, and corrected (wherever applicable) by two independent members of the research team (Stage 4). In this stage, we ensure that each scenario can be unambiguously interpreted as evoking the corresponding emotion category, and does not contain any culture- or demographic-specific references. At the end of the process, we retain 270 high-quality, rigorously validated scenarios across 15 emotion categories. To each of these scenarios, we add a question querying about each of the 17 appraisal dimensions. For example, we add to a scenario for the dimension of \textit{pleasantness}, the following: \textit{"How pleasant or unpleasant was it to be in the situation?"}. The appraisal questions are taken verbatim from~\citet{smith1985patterns}(Appendix~\ref{sec:app_example_core_benchmark}). 

\label{sec:data_collection}

\textbf{Collecting Human Responses.} To each scenario in our benchmark, we prepend the hypothetical framing ``\textit{Imagine that you recently...$<$scenario$>$}'', creating the common emotional stimuli. As we aim to obtain a behavioral representation of the underlying emotion concept, contrary to existing works~\citep{tak2025aware, tak2024gpt}, we also name the emotion directly, appending ``\textit{...and you felt $<$emotion$>$}''. Fixing the emotion concept for a given scenario ensures that variation across scenarios reflects variation within one emotion instead of disagreement about which emotion can be interpreted. 

We recruit 72 US-based participants through Prolific~\citep{Prolific2026} in an IRB-approved study. Each participant rates 30 scenarios, sampled at random across the 15 emotion categories. For every scenario, a participant rates all 17 appraisal dimensions on a scale from 1 to 11~\citep{smith1985patterns}. Participants are provided the option to skip any individual scenario they do not wish to respond to, and are compensated at \$11/hour for completing the study (additional protocol details in Appendix~\ref{sec:app_human_study_protocol}). 
After filtering out low-quality responses(Appendix~\ref{sec:app_human_study_protocol}), we retain an average of 3.7 ratings per dimension, per scenario. We assess the reliability of the data collected at four progressive and complementary levels. We find inter-rater agreement, measured through the Intra-class Correlation Coefficient (Appendix~\ref{sec:app_inter_rater}), to be modest (ICC(1,$k$) $=0.441$ at $k\approx3.7$) but significantly above chance ($p<0.05$), in line with established evidence of individual variability in human appraisals~\citep{troiano2023dimensional, peacock1990stress, cowen2017self}. Despite theoretical evidence, we empirically examine whether the observed variability stems from noise or genuine, systematic differences across individuals (Appendix~\ref{sec:app_stability_individual}). Individual differences in how scenarios are appraised are found to replicated strongly across two random, disjoint halves of the scenarios a participant rates (Pearson $r=0.66$, significantly above chance, $p<0.05$). In addition, we evaluate each rating for theoretical plausibility, using predictions of appraisal-emotion association from a large-scale meta-analytic review~\citep{yeo2024associations} (Appendix~\ref{sec:app_agreement_theory}). The human appraisals obtained in our study largely follow both the direction and the relative magnitude of theoretically predicted associations. Finally, we also validate that, across two disjoint, random halves of the human participants (partitioned 500 times randomly), the mean appraisal profiles for a given emotion are strongly correlated($r=0.911$; $95\%$ CI $[0.877, 0.934]$). This establishes that the human appraisal representations collected stem from systematically varying individuals, align with theoretical expectations, and remain reliable at the level of each emotion concept.


\textbf{Eliciting Responses from LLMs.} We primarily evaluate 8 popular LLMs with humans. We include proprietary models from two families, including relatively older and frontier models: GPT-o4-mini\footnote{https://openai.com/index/o3-o4-mini-system-card/}, GPT 5.6 Luna\footnote{https://openai.com/index/gpt-5-6/}, Gemini 2.5 Flash~\citep{comanici2025gemini}, and Gemini 3.7 Flash\footnote{https://blog.google/innovation-and-ai/models-and-research/gemini-models/introducing-gemini-3-7-flash/}, allowing comparisons of how representations change with increasing general capability. We also evaluate several open-source models, spanning a large size range, capability (general and frontier), and providers: Phi 4 14B~\citep{abdin2024phi}, Muse Glimmer 30B\footnote{https://research.meta.ai/blog/introducing-muse-glimmer-open-agentic-model}, QwQ 32B~\citep{qwq32b}, and DeepSeek R1 (671B)~\citep{guo2025deepseek}. In a subsequent analysis of post-training effects, we also study six model checkpoints from the Olmo 3 7B family~\citep{olmo2025olmo}. We query all models with the verbatim, first-person, hypothetical framing of prompts that humans receive, also directly naming the emotion concept. Models are also queried for ratings along each appraisal dimension independently to ensure structured responses that are not affected by the order of appraisal dimensions presented. Models are sampled at respective default temperatures, with "medium" reasoning effort wherever applicable. For further details on implementation, see Appendix~\ref{sec:app_llm_data}. 

\section{Results}
\subsection{Comparing Behavioral Representations of Emotion Concepts}
\label{sec:results_alignment}

Through responses obtained from both LLMs and humans on the scenarios in CoRE, we represent each agent's behavioral representation of an emotion concept in a common appraisal space. For an agent $a$ and an emotion category $e$, let $\mathcal{S}_e$ be the set of scenarios for $e$, with $n_e = |\mathcal{S}_e|$, and $d = 17$ is the number of appraisal dimensions. The behavioral representation of $e$ by $a$ is the matrix
\begin{equation}
  \mathrm{E}^{a}_{e} \in \mathbb{R}^{n_e \times d},
  \qquad
  \big[\mathrm{E}^{a}_{e}\big]_{ij} = r^{a}\!\left(s_i, d_j\right),
  \quad s_i \in \mathcal{S}_e,
\end{equation}
whose $i$-th row represents appraisal ratings provided by $a$ for the $i$-th scenario, across all $d$ dimensions. Every emotion category can therefore be represented as a set of $n_e$ points in a common $d$-dimensional appraisal space, with $\sum_{e} n_e = 270$. As a preliminary check, through Principal Component Analysis (PCA), we establish that all models and humans share a broadly similar low-dimensional geometry (Appendix~\ref{sec:app_pca}). All agents use a similar number of effective appraisal dimensions to represent emotions ($\leq6$). With coarse correspondence established, we compare the finer structure of these representations~\footnote{For each following result, we present robustness to controlling for sampling noise in the human data in the respective appendices.}.
\textbf{LLMs hold significantly less diverse and more discrete representations of emotion concepts.} We first ask how emotions from \textit{different} categories are represented within the appraisal space. Based on established ways of quantifying increased categorical perception (higher between-category separation and within-category compression)~\citep{goldstone2010categorical}, for each agent, we compute two distances: a between-emotion ($\mathrm{D_{Between}}$) and within-emotion ($\mathrm{D_{Within}}$) distance, characterizing the homogeneity of an individual emotion concept, and overall discreteness. As our design allows representing each emotion concept as a distribution, we use the multivariate Wasserstein-2 distance metric~\citep{kantorovich1960mathematical, gabriel2019computational} to compute both distances. For a given agent, $\mathrm{D_{Between}}$ is the average of all pairwise $W_2$ distances across the emotion categories. To calculate $\mathrm{D_{Within}}$, we consider the mean $W_2$ between two random halves of the representations across all scenarios within the given emotion, over $s$ random splits. For our main results, we set $s=30$ (for robustness results, see Appendix~\ref{sec:app_separation_ratio}). A high $\mathrm{D_{Between}}$ indicates that an agent clearly separates different emotion categories, while a low $\mathrm{D_{Within}}$ would indicate that the agent represents a single emotion in a homogeneous manner. The ratio of the two distances, when \textit{high}, would indicate less \textit{per-emotion diversity} and higher \textit{discretization}. 

\begin{wrapfigure}{r}{0.55\textwidth}
\centering
    \includegraphics[width=\linewidth]{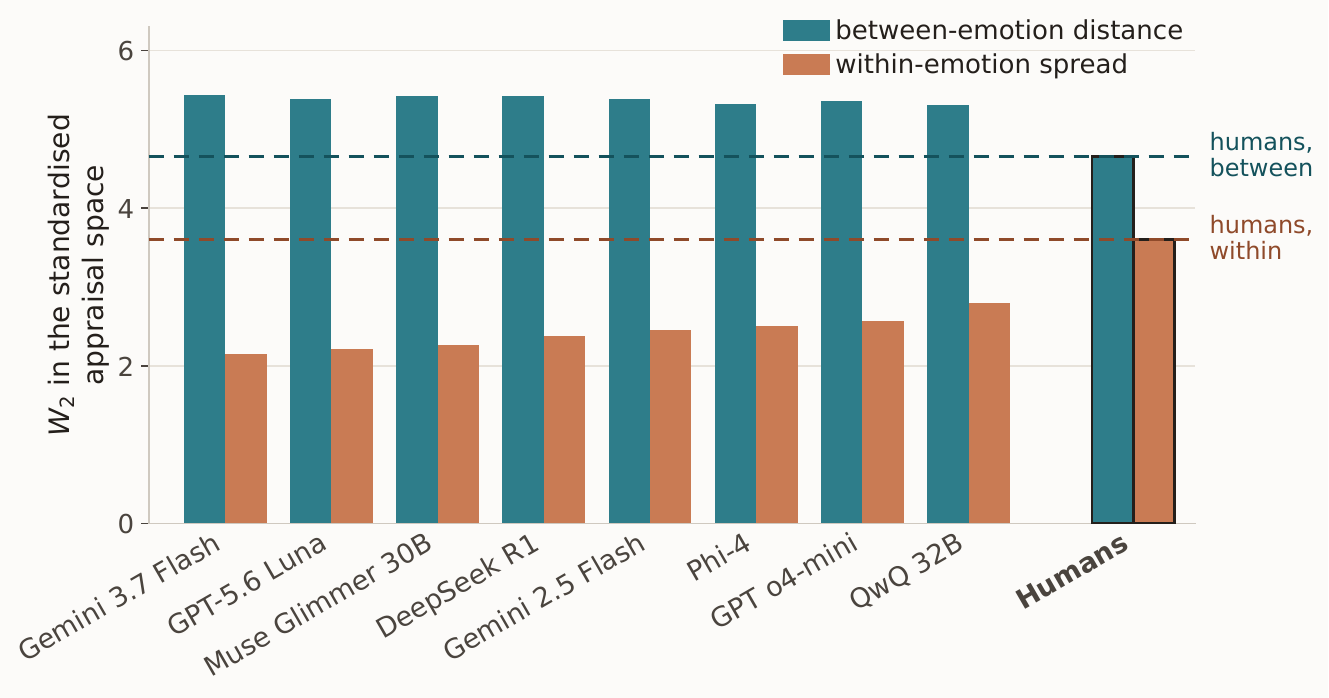}
    \caption{Between-emotion and within-emotion distance measures, computed for each agent individually, using the $W_2$ Wasserstein distance metric. Statistical significance is tested across each pair of emotions ($\mathrm{D_{Between}}$) and for each emotion category ($\mathrm{D_{Within}}$), comparing each model with humans, using a two-sided Wilcoxon signed-rank test, and rank-biserial correlation is also computed as effect sizes. Holm-Bonferroni method~\citep{holm1979simple} is used for correction of multiple comparisons, and all comparisons are significant ($p < 0.001$).}
    \label{fig:separation_ratio}
\end{wrapfigure}

We find that the separation ratio for humans is lower than \textit{all} LLMs. Further, $\mathrm{D_{Between}}$ is significantly higher for all LLMs, across all emotion pairs, while $\mathrm{D_{Within}}$ remains significantly lower for all models and each emotion category (Fig.~\ref{fig:separation_ratio}). Notably, frontier models like GPT 5.6 and Gemini 3.7 Flash, exhibit the least variation within a given emotion concept and the strongest separation between different emotion concepts (details in Appendix~\ref{sec:app_separation_ratio}). The relative lack of discretization in humans also corroborates independent evidence from human psychology that shows separation of emotion concepts in humans follows smoother gradients~\citep{keltner2023semantic}. 


\textbf{Different emotion concepts are more easily distinguishable based on LLMs' behavioral representations.} As a result of high discretization, emotion concepts are also more distinguishable in LLMs than humans. To measure this, we predict each emotion category from the scenario-specific appraisal representation, with the success of the predictor denoting separability. Using a logistic classifier as the discriminator,
we find that human appraisal representations separate emotion concepts only weakly, with a macro-F1 score of 0.26, compared to 0.63--0.80 for models (full results in Appendix~\ref{sec:app_f1}). 
The same contrast replicates when using an interpretable discriminator~\citep{kuriabov2024synthtree}. In a pairwise discrimination setting using SynthTree~\citep{kuriabov2024synthtree}, cross-validated balanced accuracy is $0.76$ for humans against $0.94$--$0.97$ across the eight models, with separability of every emotion pair higher in LLMs' behavioral representations. Further, LLMs achieve this separability using far fewer dimensions than humans: discriminators trained on LLMs' appraisal space distribute importance to an effective $2.8$--$4.0$ dimensions for a given pair of emotions, compared to $9.1$ for humans(Appendix~\ref{sec:app_synthtree}).

\textbf{LLMs are more consistent across implicit and explicit judgments of what distinguishes given emotion concepts.} Our results so far document discretization and determinacy of emotion concept representations elicited through a single format. We next test whether the same internal consistency holds for agents, across two different response formats. Specifically, for both LLMs and humans, we collect additional responses, providing an \textit{explicit} ranking of appraisal dimensions, based on how strongly, for them, it separates a given pair of emotions. Then, we correlate for each agent, this explicit ranking with the extent to which each appraisal dimension actually separates a given pair of emotions in their behavioral representations, measured directly from the behavioral representations using Cliff's $\delta$ (details of data and specific method in Appendix~\ref{sec:app_implicit_explicit}).

LLMs show moderate-to-strong consistency (mean $\rho$ up to 0.55, all significant with Holm-Bonferroni correction, at $p<<0.001$) between their explicit statements and implicit representations about which dimensions separate a given pair of emotions, whereas humans show none of this correspondence (mean $\rho = 0.06$, $p = 0.07$) (Table~\ref{tab:implicit-explicit}). Among the LLMs, the highest correspondence is again observed for the frontier-scale Gemini 3.7 Flash ($\rho = 0.55$). The same trend holds when the implicit ranking is obtained by fitting the interpretable discriminator~\citep{kuriabov2024synthtree}: the dimensions such a discriminator relies on agree with the agent's own stated discriminators at a rank correlation of up to $0.43$ for LLMs, against $0.04$ for humans (Appendix~\ref{sec:app_synthtree}). This shows that model representations of emotion concepts are not only highly discretized, but also determinate and consistent across different elicitation formats, unlike in humans. 

\begin{figure}
    \centering
    \includegraphics[width=\linewidth]{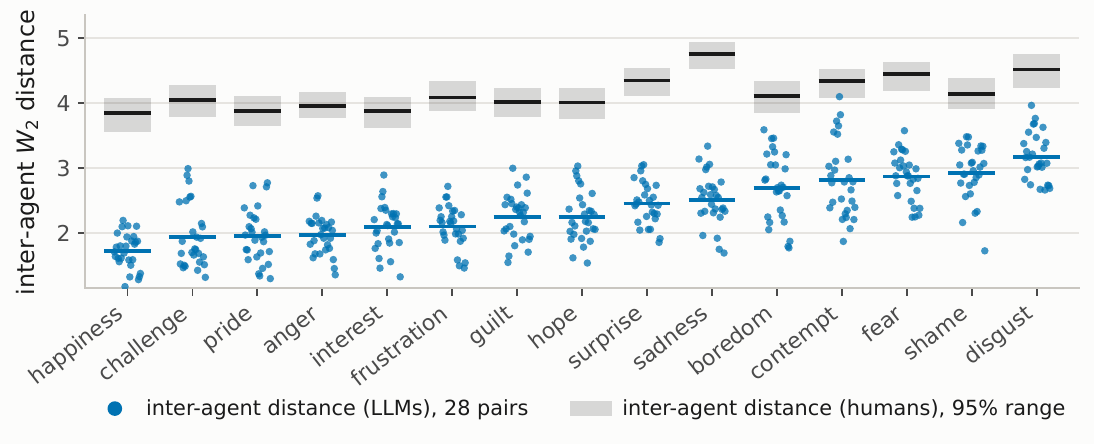}
    \caption{2-Wasserstein distance of emotion representations, comparing two different sources. For humans, to match the number of models, eight random groups are created (sampled $300$ times) with representations averaged within each group to represent a single source. The 95\% range of human distances is presented.}
    \label{fig:inter_agent_convergence}
\end{figure}


\textbf{Different LLMs agree on an emotion representation more than different humans.} Our analyses so far establish through complementary analyses that each LLM \textit{individually} produces more discrete and determinate behavioral representations of emotions compared to humans. We next ask whether this agent-specific concept-level homogeneity extends across multiple LLMs. Specifically, we investigate whether different LLMs concur on the appraisal-based representation of a given emotion category more than different groups of humans. We define agreement/concurrence in a relative sense using the multivariate $W_2$ distance between behavioral representations of the \textit{same} emotion concept from \textit{different} LLMs or groups of people (details in Appendix~\ref{sec:app_convergence}). 

For each of the 15 emotion categories, we find that LLM-LLM agreement is significantly higher (lower distance) (Wilcoxon's signed-rank test, $p < 0.001$) than agreement between any two random groups of the human raters (Fig.~\ref{fig:inter_agent_convergence}). This denotes that the consistency of behavioral emotion representations extend beyond a single artificial agent, even beyond models within the same family. Notably, we do not detect a significantly higher agreement between models from the \textit{same} families. This suggests that, more than the shared training mechanisms for models within a family, the globally shared data models are pretrained on could be driving the concurrence. We also find that the heightened inter-agent similarity for LLMs hold when using a distance-free measure to quantify concurrence in Appendix~\ref{sec:app_convergence}.

\begin{figure}[t!]
    \centering
    \includegraphics[width=0.9\linewidth]{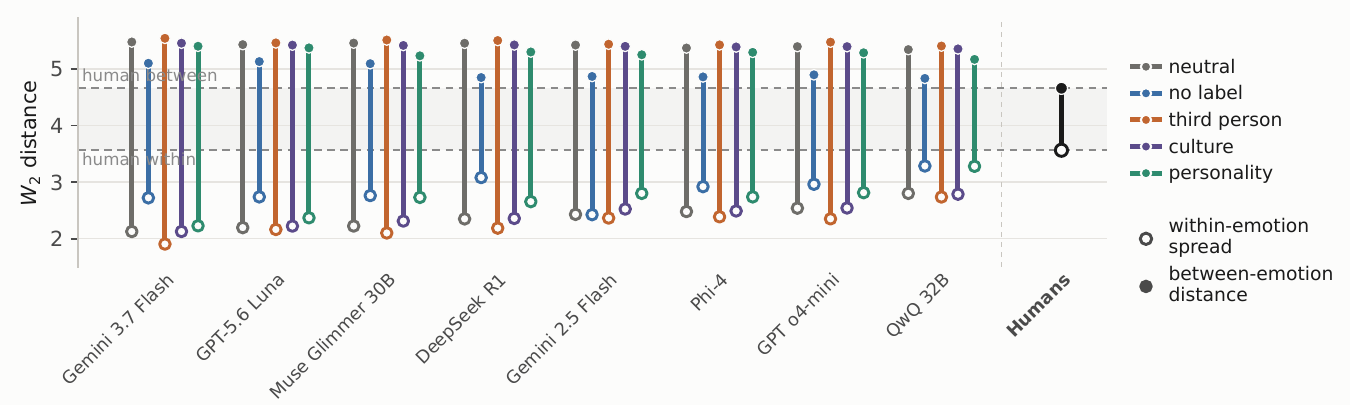}
    \caption{Effect of: (blue) removing the emotion label from the prompt, (red) eliciting responses with a third-person framing (purple) inducing nationality-based cultural personas, and (green) conditioning on personality traits on the separation ratio and its distance components.}
    \label{fig:context_main}
\end{figure}

\subsection{Varying Prompt Context and Post-Training Strategies}
\label{sec:results_ablation}



\textbf{Eliciting Responses without an emotion label from LLMs.} A natural question that emerges from our results is: do LLMs represent emotion concepts with lower diversity only when the concept term is directly made available in the prompt? We thus replicate our main experiment of eliciting appraisal ratings with all of our models, without providing the emotion concept directly in the prompts. As shown in Fig.~\ref{fig:context_main}, we find that removing the emotion label from the prompt reduces between-emotion separation and increases within-emotion separation to some extent for all LLMs. However, every LLM remains significantly more discretized (Wilcoxon signed-rank test, all $p<0.001$) and less internally diverse ($p < 0.05$) than the human representations, where the human raters see the emotion concept directly. Noting that not providing the emotion label in the prompt could change the construal of the emotion concept itself, we use a separate LLM-judge to map free-form model responses to one of the 15 emotion categories (Appendix~\ref{sec:app_without_emotion}). Indeed, we find that models misassign emotion concepts 47\% of the time, on average. Even considering only samples where LLMs correctly interpret the emotion concept from the scenario, we find that their representations remain significantly more discretized than humans, ruling out the possibility that the categorical nature of behavioral emotion representations in LLMs is an effect of prompting directly by naming the emotion concept.

\textbf{Varying Task Framing and Personas.} We consider three different contextual changes: (a) changing the task framing by prompting models to produce a third-person appraisal (details of setup in App.~\ref{sec:app_third_person}), (b) including nationality-based cultural personas~\citep{alkhamissi2024investigating, kamruzzaman2025anger} (Appendix~\ref{sec:app_culture}), and (c) including personas based on Big-5 trait descriptors~\citep{liu2025synthetic} (Appendix~\ref{sec:app_personality}). As seen in Fig.~\ref{fig:context_main}, third-person framing further increases discretization and lowers diversity of emotion representations. We also find that emotion representations change qualitatively under third-person framing (full results in Appendix~\ref{sec:app_third_person}), with models consistently appraising the same negatively valenced scenarios as less effortful and aversive with the third-person framing.

\begin{figure}
    \centering
    \includegraphics[width=\linewidth]{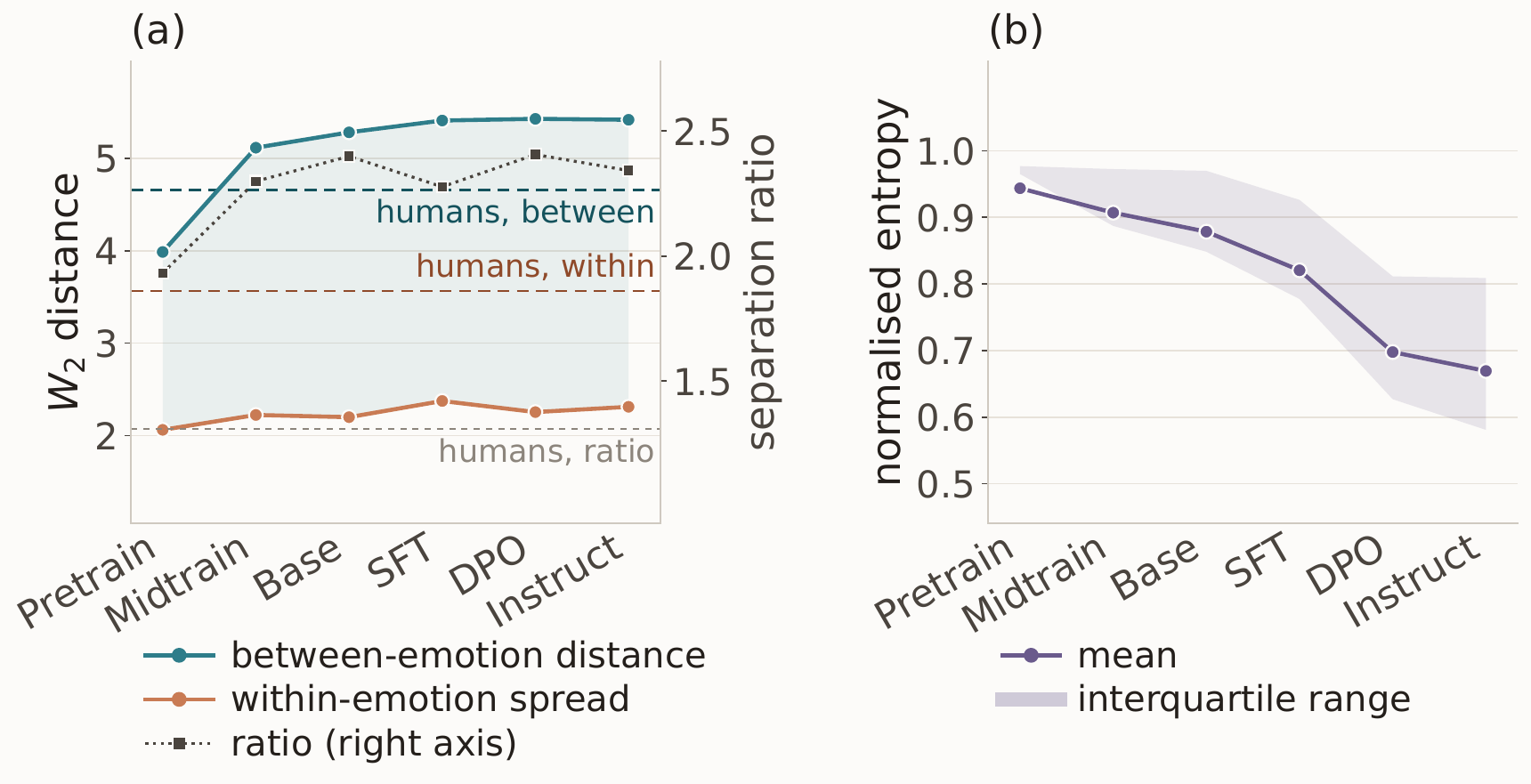}
    \caption{(a) Separation ratio across the Olmo model checkpoints, across different training stages, (b) Normalised entropy of the probability distribution over the rating scale at the answer position.}
    \label{fig:olmo_ladder}
\end{figure}


The structure of emotion representations with nationality-based cultural framing, however, remains essentially identical to the neutral, persona-free condition for every model (Fig.~\ref{fig:context_main}), with both the between- and within-emotion separation remaining significantly different from humans ($p<0.01$ for both quantities). Finally, personality-based personas do lead to a significant reduction in the separation ratio for all models (Fig.~\ref{fig:context_main}), driven particularly by change in within-emotion spread. However, $\mathrm{D_Between}$ for \textit{all} model-personality combinations remain significantly higher than for humans, while $\mathrm{D_Within}$ for all models other than QwQ and Gemini 2.5 Flash remain significantly lower. On average, the effect of personality traits is strongest for neuroticism, followed by agreeableness, conscientiousness, openness, and extraversion. We, however, note that the personas that specifically lead to the strongest reduction in the separation ratio also make representations more \textit{directionally} misaligned with the human distribution and theoretical predictions (Appendix~\ref{sec:app_personality}). This suggests that while extreme personality-based personas could increase variability of appraisal-based representations, it is at the cost of being directionally aligned with a general human population~\citep{xiao2026chameleon}. Thus, with most contextual variations, even though behavioral representations of emotions from LLMs shift qualitatively, their homogeneous and discretized nature does not change significantly.

\textbf{Tracing discretization across training checkpoints.}
\label{sec:post_training}
Beyond contextual changes in the prompt, we probe the effect of post-training, using model checkpoints at different training stages from the OLMo 3 family~\citep{olmo2025olmo} (details of setup in App.~\ref{sec:app_post_training}). Plotting how the quantities composing the separation ratio change across different training stages (Fig.~\ref{fig:olmo_ladder}(a)), we find that within-emotion diversity is low at all stages of training, remaining significantly lower than the human representations. This suggests that within-concept homogeneity stems potentially from how internet-scale data is learned in the next-token-prediction paradigm. Discretization of emotion concepts, on the other hand, remains lower than even humans at pretraining, and changes the most after midtraining on reasoning-heavy data. After the base model is trained, different alignment strategies do not meaningfully change these quantities. Examining the probability distribution at the answer position, however, we find that post-training leads models to become more decisive about the final \textit{output}, without changing behavioral representations at the \textit{concept} level (Fig.~\ref{fig:olmo_ladder}(b)).

\section{Discussion}
Recent research argues that LLMs merely represent emotions to aid context-sensitive interpretation of situations, and differ from humans in how emotions functionally influence internal processing~\citep{goldenberg2026large}. Even when only representing emotions to interpret situations, our results show how LLMs markedly differ from humans, reducing the variability constitutive of emotions in humans~\citep{siegel2018emotion, barrett2009variety}. This holds true even when the comparison is with a human distribution relatively homogeneous in terms of demographics (US-based participants). Further, as we show, even when contextual variations qualitatively shift the appraisal representations, their discretized nature does not change. 

This determinacy of emotion representations in models can in many ways be both an asset and a deficiency. Repetitive tasks, such as reproducible annotation of affective content, can benefit from the lack of variability in LLMs, that human raters may not provide. Yet, in affective tasks requiring context-specific, pluralistic judgment, such as simulating a diverse human population~\citep{bhattacharyya2026expressing, wang2025user, hu2026simbench}, it could be a limitation. The homogeneous, convergent nature of emotion representations could further adversely impact user-facing interactions with LLMs involving emotional support or advice~\citep{phang2025investigating, rousmaniere2026large}. When a user reports feeling guilt, for example, a model that represents guilt through a single typical appraisal pattern will provide feedback that may only be \textit{generally} applicable, even when the specific scenario may lead to distinct pattern of emotional interpretation for a user. Repeated and omnipresent use of AI systems that represent inherently variable concepts like emotions in a determinate and convergent manner could exacerbate harms like reduced diversity of collective human thought~\citep{doshi2024generative, sourati2026homogenizing}. 

Beyond specific downstream effects, our results point to a more fundamental misalignment. In humans, variability in how emotions are understood enables adaptation to individual and contextual circumstances, supports emotion regulation, and serves social functions~\citep{smidt2015brief}. In LLMs (machines that do not have goals of their own~\citep{quattrociocchi2025epistemological}) the incentive for retaining diverse representations of emotions is absent. As our analyses on the Olmo checkpoints suggest, after pretraining with internet-scale data itself, models learn internally homogeneous representations of abstract emotion concepts. This is distinct from the reduced diversity of \textit{outputs} due to post-training~\citep{kirk2024understanding}, reported in existing literature on LLM mode collapse, where homogeneity is measured at the level of individual outputs to a single prompt~\citep{jiang2025artificial}, and not a concept. At the level of a concept, the pattern we highlight is not general mode collapse, since \textit{distinct} emotion concepts remain more separated while the variation within each concept is lost. This parallels the finding from~\citet{shani2026from} where LLMs trade off semantic richness for efficient compression of concepts more objective than emotions (e.g., bird, furniture, etc.). 

Our findings thus highlight \textit{concept}-level homogeneity and discretization as a fundamental form of structural difference in how LLMs and humans behaviorally represent emotion concepts. We hope that the release of our benchmark and evaluation methods can motivate further nuanced evaluation of LLM behavior, recognizing that agreement with humans on individual outputs does not ensure agreement in how broader, abstract, and subjective concepts are structured.

\subsection*{AI use statement}

In this work, we used generative AI tools to generate synthetic datasets, design or provide feedback on research  methodology or experiments, implement methods. We have not used generative AI tools to
help develop theoretical models or conceptual frameworks, propose or refine hypotheses, clean and reformat dataset, interpret results, and formulating mathematical claims, providing critical ingredients for proving mathematical claims, assisting in the writing of proofs, assisting with translation, and supporting qualitative and thematic data analysis are not applicable to this work. Additionally, we used generative AI tools to create or modify scientific figures or images, create or edit software code, draft parts of a research paper, brainstorm, and propose a title or keywords for a research paper. For all of the declared uses, the Claude Opus 5 and 5.5 models were used, through private chat-based subscriptions.

We have reviewed all AI-assisted work. The AI-assisted data generation process is verified through two additional, independent rounds of human validation. In each round, each AI-generated data sample was verified, corrected, rejected, or rephrased by two independent individuals. All AI-generated code was verified for correctness by one author, through multiple complementary tests and cross-checking. All figures created or modified by AI were re-verified according to the actual data and results by two authors. No AI-generated drafts of the paper are directly used in the paper, without thorough review, editing, and changes. We take responsibility for the final content of this work,
including text, claims or artifacts produced with the aid of generative AI.

\subsection*{Ethics statement}

All human data collection was reviewed by the Institutional Review Board of the authors' institution and determined to be exempt. Participants were adults in the United States who gave informed consent, could withdraw at any point or skip any question, and were compensated for their time. Scenarios were restricted to everyday situations, and the collected data contain no directly identifying
information. As part of the benchmark release, human responses will be released in de-identified form. Our findings concern behavioral representations of emotion elicited through appraisal judgments, and make no claim that LLMs experience emotions. Because the human sample is drawn from a single country, our conclusions do not describe human emotion universally. The authors declare no competing interests.

\subsection*{Reproducibility statement}

The construction and validation of the benchmark are described in Section~\ref{sec:benchmark} with details in Appendix~\ref{sec:app_core_benchmark}, and verbatim examples in Appendix~\ref{sec:app_example_core_benchmark}. The human data collection protocol, including exclusion
criteria, compensation details and protocol details are described in detail in Appendix~\ref{sec:app_human_study_protocol}.
The exact prompt templates used for humans and for models are given in Appendix~\ref{sec:app_prompt_templates}. Model implementation details are provided in Appendix~\ref{sec:app_llm_data} for the main experiments and in Appendix~\ref{sec:app_post_training} for the Olmo 3 checkpoint analyses. All analysis measures and statistical tests are defined in Section~\ref{sec:results_alignment}, and the robustness analyses, including the
measurement-matched controls, are reported in the Appendices~\ref{sec:app_pca} through~\ref{sec:app_personality}. The anonymized code and the benchmark scenarios are provided as supplementary material, while we plan to release the complete data set upon publication. 



\subsection*{Acknowledgments}
This material is based upon work supported by the National Science Foundation (NSF) under Award No. 2234195 and by scholarships awarded to S.B. from Penn State's J. Jeffrey and Ann Marie Fox Graduate School. This work used computing resources at the National Center for Supercomputing Applications and the Pittsburgh Supercomputing Center through an allocation from the Advanced Cyberinfrastructure Coordination Ecosystem: Services \& Support (ACCESS) program. ACCESS is supported by NSF Award Nos. 2138259, 2138286, 2138307, 2137603, and 2138296. We thank the anonymous reviewers and editors for their careful and constructive comments, which helped improve the manuscript.

\bibliography{iclr2027_conference}
\bibliographystyle{iclr2027_conference}

\appendix
\clearpage
\startcontents[appendices]
\section*{Appendix}
\printcontents[appendices]{}{1}{\setcounter{tocdepth}{2}}

\section{Discussions}

\subsection{Limitations}

Several limitations apply to our current work, and we describe three key limitations below. 

\paragraph{Contextual and individual variability.} The goal of our study is to assess the structural similarity of emotion representations in humans and LLMs. In doing so, our primary focus remains on the variability of an emotion concept across situations: for each scenario, human ratings are averaged across participants, so that the human representation of an emotion reflects how its appraisal differs from one situation to another. Averaging partially eliminates a second source of variability, the differences between individuals who appraise the same situation, which prior work has studied in a third-person appraisal setting~\citep{zhou2025modeling}. Our dataset and evaluation methodology also support the analysis of this second source, since each scenario is rated by several participants and individual ratings are retained. Our analytical conclusions therefore concern variability across situations, and do not extend to variability across individuals.

\paragraph{Operationalization of behavioral representations.} We characterize emotion concepts through behavioral representations, as described in
Section~\ref{sec:introduction}, because these can be elicited in a comparable manner from humans
and from LLMs. We operationalize these representations through cognitive appraisal
theory~\citep{smith1985patterns}, chosen because LLMs have been shown to produce reliable appraisal judgments that are comparable to human judgments at the level of individual outputs~\citep{tak2024gpt, gandhi2024human}. Cognitive appraisal is, however, one of several theoretically grounded ways to characterize emotion. Concurrent frameworks include dimensional theories, which characterize an emotion by core affect, principally valence and arousal~\citep{russell2003core}, action-readiness theories, which characterize an emotion by the tendencies toward action that it produces~\citep{frijda1989relations}, and semantic space theory, which locates an emotion in a high-dimensional space of reported emotional experience~\citep{cowen2021semantic}. We do not examine these alternatives in the current study, and
the degree of structural alignment we report may depend on the dimensions used to characterize an emotion.

\paragraph{Diversity of the human sample.} Our human data are collected from a relatively small pool of participants residing in the United States. A pool of this size and composition is sufficient for an evaluation-scale dataset, but would not be sufficient for a dataset intended for training. A more diverse sample would be expected to increase the variability observed in human appraisals, since people from different cultural backgrounds appraise the same situations differently~\citep{scherer1997role, mesquita2001role}. The homogeneity of our sample
therefore makes the comparison stricter: the human reference is less variable than a more diverse sample would be, and a more diverse reference would be expected to make LLM representations appear more discretized in comparison. We do not test this expectation directly in our current work, and our findings describe alignment with a single, demographically restricted population.

\subsection{Future Directions}

\textbf{Further evaluations.} Subjectivity in interpreting or perceiving emotion concepts can be considered to have at least two independent sources of variability: context-specific and individual-specific. While some studies have focused on individual variability (through third-person inter-rater disagreement~\citep{zhou2025modeling}, for example), we focus on measuring subjectivity stemming from contextual variability, i.e., variability across situations within an emotion concept. We operationalize this through averaging multiple human ratings to make the ``individual" constant. The complementary component of subjectivity, how differently two individuals appraise the same situation, and the ensuing structure of individual disagreement, can also be measured using our dataset and evaluation methods.
The same methodology could also apply to other domains in which human judgments legitimately vary, such as moral and aesthetic evaluation, where existing work has largely focused so far on comparing humans and models on whether their output-level judgments agree~\citep{hosseini2026judgment, abe2026interoceptive}, instead of whether the underlying concepts are organized similarly.

\textbf{Interventions at the data layer.} Pretraining corpora contain descriptions of emotional situations written by many authors, and descriptions of a given emotion converge on its typical eliciting conditions. Collecting theory-grounded judgments at a scale suitable for training could therefore be a valuable future direction. A larger-scale dataset that elicits human judgments for the same set of emotion-laden scenarios, across demographically diverse populations, could provide training signal that captures both forms of subjectivity. In parallel, web-scale unsupervised resources could also be leveraged to curate high-quality data on emotion judgments, following success observed in other subjective tasks like predicting visual memorability~\citep{bhattacharyya2026unsupervised}. 

\textbf{Interventions at the training layer.} All models we evaluate in our study are language models trained to predict the next token, and the behavioral representations we measure are learned as a by-product of that objective. Training paradigms that operate over larger semantic units offer an alternative: large concept models predict in a sentence embedding space rather than over tokens~\citep{barrault2024large}, and concept-aware training objectives make conceptual structure an explicit target~\citep{iyer2026beyond}. Whether such
alternate training objectives change how subjective concepts are internally organized is an open question, and the measures introduced here provide a direct way to test it by comparing the within-concept and between-concept structure of models trained under different objectives. 

\section{Methodological Details for the CoRE Benchmark}
\label{sec:app_core_benchmark}

\subsection{Details of Appraisals and Emotions}
\label{sec:app_appraisal_emotion}

\begin{table}[t]
\centering
\small
\setlength{\tabcolsep}{4.5pt}
\resizebox{\textwidth}{!}{
\begin{tabular}{@{}lcccccccc@{}}
\toprule
          & anger & frustration & contempt & fear & disgust & boredom & sadness & guilt \\
\midrule
Scenarios & 22 & 16 & 17 & 20 & 17 & 17 & 16 & 18 \\
Valence   & neg. & neg. & neg. & neg. & neg. & neg. & neg. & neg. \\
\midrule
          & shame & challenge & surprise & hope & interest & pride & happiness & \\
\midrule
Scenarios & 18 & 25 & 17 & 18 & 20 & 15 & 14 & \\
Valence   & neg. & amb. & amb. & pos. & pos. & pos. & pos. & \\
\bottomrule
\end{tabular}}
\caption{The fifteen emotion categories and the number of scenarios rated for each, over the $270$ scenarios in the benchmark.}
\label{tab:app_emotions}
\end{table}

\begin{table}[t]
\centering
\small
\begin{tabular}{@{}lll@{}}
\toprule
Family & Dimensions & Scale as presented \\
\midrule
Valence          & pleasantness, enjoyment   & $-5$ to $5$\\
Attention        & attention, consideration  & $-5$ to $5$\\
Comprehension    & understand, certainty     & $1$--$11$ \\
Control          & self, other, situational  & $1$--$11$ \\
Goal obstruction & problem, obstacles        & $1$--$11$ \\
Legitimacy       & cheated, fairness         & $1$--$11$ \\
Responsibility   & self, other               & $1$--$11$ \\
Effort           & exert, effort             & $1$--$11$ \\
\bottomrule
\end{tabular}
\caption{The seventeen appraisal dimensions, grouped by what each question asks about. Every dimension is a single rated question. Pleasantness and attention were presented on a $-5$ to $5$ scale, following the original study~\citep{smith1985patterns} and shifted to $1$--$11$ for
analysis, so that all seventeen dimensions share one scale.}
\label{tab:app_appraisals}
\end{table}

We provide a complete list of the 15 emotion categories that we use in our study, drawing directly from~\citet{smith1985patterns}, in Table~\ref{tab:app_emotions}. Along with that, we also show the number of scenarios included in our benchmark for each emotion category. We also provide a list of the 17 appraisal dimensions we use in our study in Table~\ref{tab:app_appraisals}. We provide brief definitions of the broad categories of appraisals here, drawing on the descriptions provided in the original human study~\citep{smith1985patterns}:

\begin{itemize}
    \item \textit{Attentional Activity:} Cognitive appraisal dimensions under this broad category represent an ``orienting" appraisal, where an agent determines whether to attend to a given stimulus, ignore it, or avoid it~\citep{scherer1982emotion}. It is often also considered the first appraisal necessary for all other subjective appraisals of a given situation to follow. Emotions situated at opposite poles of attentional activity might include \textit{interest} (strong allocation of attention to the given scenario) and \textit{disgust} (tendency to avoid the stimulus causing the emotion). 
    \item \textit{Certainty:} Dimensions of cognitive appraisal under this broad category represent how well an agent understands what is going on in the situation. For example, \textit{hope} and \textit{fear} can be considered emotions evoked in situations where a low level of certainty is involved. 
    \item \textit{Control:} Control primarily represents how well an agent evaluates their coping ability to be, in a given situation. Similar to the existing human study~\citep{smith1985patterns}, we include the two \textit{distinct} types of control dimensions: by a person (self- or other-control) and by circumstance. 
    \item \textit{Pleasantness:} An appraisal dimension that arises dominantly in several distinct theories of appraisal~\citep{scherer1982emotion}, pleasantness can be construed in at least two distinct ways. It can represent the view that any given stimulus has an intrinsic valence/pleasantness associated with it. On the other hand, it can also be construed as a combination of the appraisal of what one wants in relation to what one has (goal-achieving). Keeping with the original human study~\citep{smith1985patterns}, our paper operationalizes the former definition. 
    \item \textit{Perceived Obstacle:} The cognitive dimensions within perceived obstacle represent whether an agent evaluates the situation as presenting any kind of challenge, standing in the way of some desired outcome (even when the agent is unconscious of the outcome). This can also be considered semantically similar to the primary goal-based appraisal~\citep{lazarus1984stress}. Similar to the definition adopted by~\citet{smith1985patterns}, we note here that perceived obstacle need not necessarily tie to positive valence, as it could also be associated with desirable intellectual (e.g., in solving a puzzle) or physical challenge (e.g., while playing sports). 
    \item \textit{Responsibility:} This primarily involves who, according to the agents' evaluation, is responsible for initiating the situation, or the emotional cause. This could involve attributing responsibility to self or any other entity. Note that this is not necessarily the same as the control dimensions, as an entity could be responsible for initiating the situation, but then cease to control it. For example, self-conscious emotions such as \textit{guilt} or \textit{shame} often arise in situations involving high self-responsibility.
    \item \textit{Legitimacy:} This appraisal dimension concerns an agent's evaluation of whether the given outcome was deserved or undeserved. For example, the emotion of \textit{anger} is often associated with situations perceived as unfair or negative outcomes perceived as undeserving. 
    \item \textit{Anticipated Effort:} Appraisal dimensions within this category pertain to an agents' evaluation of whether they need to expend effort or do something in a situation (e.g., in classic fight-or-flight situations) or they evaluate it to be relaxing and calm. As described in the original human study~\citep{smith1985patterns}, anticipated effort can be considered related to the level of activation involved, and is physiologically very important. 
\end{itemize}

\subsection{Details of multi-rater verification of candidate scenarios}
\label{sec:app_multi_rater_verification}

\begin{figure}
    \centering
    \includegraphics[width=0.9\linewidth]{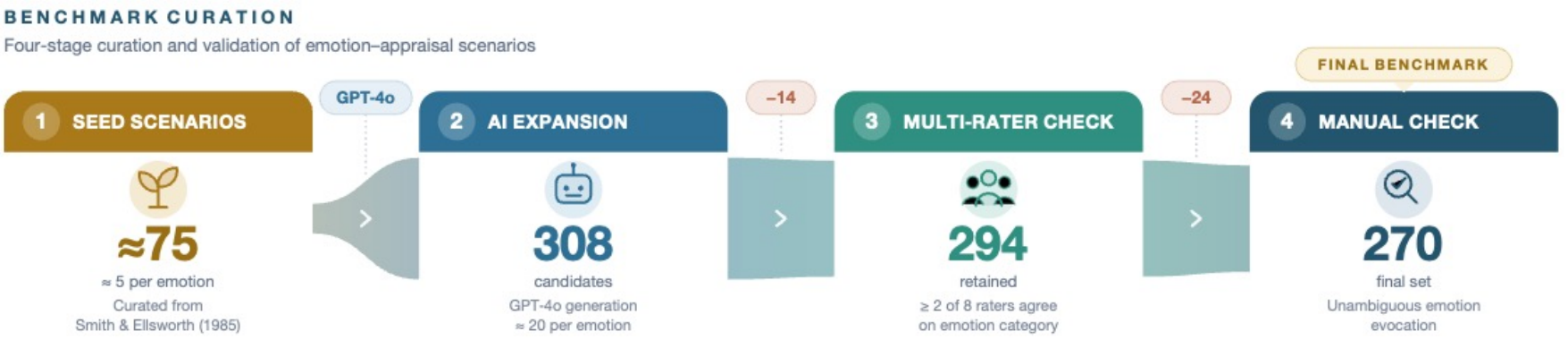}
    \caption{Overview of the dataset curation process, including the four independent stages.}
    \label{fig:app_core_curation}
\end{figure}

In this section, we provide additional details of the multi-rater verification stage in the benchmark curation (the full 4-stage pipeline is additionally shown in Fig.~\ref{fig:app_core_curation}). As stated in the main text, the key reasons for involving human annotations to validate the benchmark are: (a) to ensure that scenarios generated and included in the dataset are clearly related to the corresponding emotion categories, (b) to mitigate exclusive reliance on synthetically scaled scenarios, and (c) to ensure the scenarios are valid real-life descriptions. 

The human raters are tasked with responding to a simple question about each of the initial 308 scenarios: whether it is appropriately related to its corresponding emotion class. If the rater judges that it is related and accurately reflects the emotion, they are allowed to simply indicate this and proceed to the next scenario. If the rater, however, judges that the scenario is not well related to the corresponding emotion (e.g., it describes an unrealistic or extreme scenario, or allows an ambiguous interpretation with respect to the given emotion), they are given the option to enter an alternative scenario from daily life that more clearly reflects the given emotion category. 

Each scenario is rated by two human annotators. For each scenario, if both raters judge it to be correct, it is retained directly in its original form. If at least one of the raters provides an alternative scenario, that alternative is instead considered for inclusion in the dataset. In cases where both raters provide alternatives, both new scenarios are considered for inclusion. In this process, we obtain two independent ratings on each of the scenarios. Unanimous agreement occurs for 272 out of 308 scenarios (\(\sim\) 88\%), where the original scenarios are directly retained. Within the remaining scenarios, 28 scenarios (9\% of total) are considered valid by at least one rater, whereas the remaining 8 are rejected by both raters. Across the 36 scenarios where at least one rater rejects the scenario, we obtain 22 augmented/new scenarios provided by the raters through the optional entry platform. This leads to the final 294 scenarios retained after this step. 

Given the relatively smaller number of candidate scenarios (308), we recruit only domain-informed human raters, instead of large-scale crowdsourcing. Specifically, we recruit a total of eight participants, of whom two are senior undergraduate students in Computer Science, and six are graduate students in the fields of Computer and Information Sciences. Each task batch involved validating 20 scenarios and took roughly 10 minutes. The platform we utilized was Gorilla.sc, which provided ways to randomize the assignments made to raters to ensure responses could not be traced back to a single participant. The raters were compensated at the rate of \$12/hour. An institutional review board determined that the protocol was exempt.



\subsection{Complete Examples from CoRE Benchmark}
\label{sec:app_example_core_benchmark}

In our benchmark, we append to each scenario description, a question each querying about an appraisal dimension. Here, we provide the exact questions used for each appraisal dimensions, borrowed verbatim from the original human study we draw inspiration from~\citep{smith1985patterns}. 

\begin{itemize}
    \item Pleasantness: \textit{How pleasant or unpleasant was it to be in the situation? Answer with a text explanation, and a number from -5 to 5, with -5 for extremely unpleasant and 5 for extremely pleasant. }
    \item Enjoyment: \textit{How enjoyable was it to be in the situation? Answer with a text explanation, and a number from 1-11, with 1 for not at all enjoyable and 11 for extremely enjoyable.}
    \item Consideration: \textit{What was causing you to feel the said emotion in this situation? Did you try pursuing the cause of your emotion further, or did you try shutting it out? Answer with a text explanation, and a number from 1-11, with 1 denoting if you tried actively considering the cause further, and 11 denoting if you actively tried shutting off the cause.}
    \item Attention: \textit{Think about what was causing you to feel the said emotion in this situation. When you were feeling the said emotion, to what extent did you try to devote your attention to this thing, or divert your attention from it? Answer with a text explanation, and a number between -5 and 5, with -5 denoting strongly diverted attention, and 5 denoting strongly devoted attention.} \footnote{Note that attention and consideration are operationalized in an opposite manner.}
    \item Control-situational: \textit{ To what extent did you feel that circumstances beyond anyone's control were controlling what was happening in the situation? Answer with a text explanation, and a number from 1-11, with 1 denoting that it was not at all the case, and 11 if you think it was absolutely the case.}
    \item Control-self: \textit{To what extent did you feel that you had the ability to influence what was happening in the situation? Answer with a text explanation, and a number between 1-11, with 1 if you think you had no ability to influence the situation, and with 11 representing that you had strong ability to influence the situation.}
    \item Control-other: \textit{When you were feeling the said emotion, to what extent did you feel that someone else other than you was controlling the situation? Answer with a text explanation, and a number between 1-11, with 1 denoting that it was not at all in control of someone else other than you, and 11 denoting that you felt the situation was entirely being controlled by someone else other than you.}
    \item Certainty: \textit{When you were feeling the said emotion, to what extent were you uncertain or certain of what was happening in the situation? Answer with a text explanation, and a number between 1-11, with 1 denoting that you were absolutely uncertain of what was happening, and 11 denoting that you were absolutely certain of what was happening in the situation.}
    \item Problem: \textit{ Think about what you wanted when you felt the said emotion in this situation. While you were feeling the said emotion, to what extent did you feel there were problems that had to be solved before you could get what you wanted? Answer with a text explanation, and a number between 1-11, with 1 denoting that you felt there were no problems to be solved before you could get what you wanted, and 11 denoting that you felt there were too many problems to be solved before you got what you wanted.}
    \item Obstacle: \textit{Think about what you wanted when you felt the said emotion in this situation. When you were feeling the said emotion, to what extent did you feel there were obstacles standing in the path between you and getting what you wanted? Answer with a text explanation, and a number between 1-11, with 1 denoting that you felt that there were no obstacles in the path between you and getting what you wanted and 11 denoting that there were extremely challenging obstacles.}
    \item Legitimacy-cheated: \textit{ When you were feeling the said emotion in this situation, to what extent did you feel cheated or wronged? Answer with a text explanation, and a number between 1-11, with 1 denoting that you did not feel cheated or wronged at all, and 11 denoting that you felt extremely cheated or wronged.}
    \item Legitimacy-fair: \textit{What caused you to feel the said emotion in this situation? How fair was the cause? Answer with a text explanation and a number between 1-11, with 1 denoting that you felt that the cause was extremely unfair, and 11 denoting that you felt the cause was absolutely fair.}
    \item Responsibility-self: \textit{When you were feeling the said emotion, how responsible did you feel for having brought about the events that were making you feel the emotion in this situation? Answer with a text explanation and a number between 1-11, with 1 denoting that you felt you were not responsible at all, and 11 denoting that you were the most responsible.}
    \item Responsibility-other: \textit{ When you were feeling the said emotion, how responsible did you think someone or something other than yourself was for having brought about the events in this situation? Answer with a text explanation and a number between 1-11, with 1 denoting that you felt no external entity was responsible, and 11 denoting that some external entity or entities were entirely responsible.}
    \item Exert: \textit{ When you were feeling the said emotion, to what extent did you feel that you needed to exert yourself to deal with this situation? Answer with a text explanation and a number between 1-11, with 1 denoting that you did not need exert yourself at all, and 11 denoting that you needed to exert yourself a great deal.}
    \item Effort: \textit{When you were feeling the said emotion, how much effort (mental or physical) did you feel this situation required you to expend? Answer with a text explanation and a number between 1-11, with 1 denoting that you did not need to expend any effort at all, and 11 denoting that you needed to expend significant effort.}
\end{itemize}


\section{Human Data Collection}
\label{sec:app_human_data}

In this section, we provide additional details about the human data collection, including details on the study protocol and implementation, and measurements of reliability of the human data. 

\subsection{Study Protocol Details}
\label{sec:app_human_study_protocol}

Our human data collection is conducted through the Prolific platform, while we present the scenarios in a text-based survey, hosted through an institution-specific version of Qualtrics~\citep{Qualtrics2026}. Because the study is presented in English and requires understanding the emotion categories as described in English, we restrict recruitment to participants in the United States and pre-screen for English proficiency. A geographically homogeneous pool of participants is also chosen to ensure that reported variability of emotion representations is not confounded with variability across cultures. Participants are also required to be 18 years or older. As stated in the main body, each participant rates 30 individual scenarios, providing a rating between 1--11 for each of the 17 appraisal dimensions. We chose 30 scenarios per participant considering sustaining their engagement across the session, to keep compensation fair for the time required, and to balance the number of participants recruited against the number of ratings collected per scenario. The ratings are obtained using a slider ranging from 1--11, following the original human study~\citep{smith1985patterns}.

The maximum allowed time for completion of a single session (30 scenarios) is set to 110 minutes, based on a pilot study with 10 participants, where the median completion time was found to be 60 minutes. To ensure high quality of the data collected, each participant also receives two attention check questions, along with the 30 scenarios. An attention check presents a dummy scenario that is not part of our benchmark, with a dummy emotion category such as "calm", and instructs the participant to give the same rating on every appraisal dimension (e.g., 4 or 8). For each scenario, participants are also provided with an additional slider, which they can place anywhere above 5 to indicate that they wish to skip the scenario. 

After the data collection process is complete, we apply several post-processing steps to filter out low-quality data directly at the participant level. For example, any participant who provides ratings for less than 80\% of the total scenarios (either by leaving sliders at their default positions for all dimensions, or explicitly opting to skip) is excluded. Further, any participant failing any of the attention checks is also excluded. Filtering out data at the participant level ensures that any downstream variability stems from genuine variation across individually reliable raters. In addition to this, at the level of each scenario, we only include ratings that have the complete 17-dimensional appraisal profile, leading to the final data retaining 41 effective participants. 

\subsection{Average Inter-Rater Reliability}
\label{sec:app_inter_rater}

Because each participant rated a random subset of 30 scenarios, every scenario was judged by a \textit{different}, \textit{randomly drawn} set of raters. Averaged over the 17 appraisal dimensions, we find ICC(1,$k$) $=0.441$ at $k\approx3.7$ raters per scenario, indicating modest agreement between individual raters and moderate reliability of the averaged ratings used in our analyses. The modest agreement we report is in alignment with existing evidence of inter-participant variability in responding to the same emotional stimuli~\citep{cowen2017self}. Against a permutation-null in which ratings are shuffled across scenarios, mean ICC(1,$k$) is $-0.009$ (95th percentile $0.032$). Inter-rater reliability is thus is significantly above chance ($p<<0.05$) despite its modest absolute value.

\subsection{Stability of Individual Rater}
\label{sec:app_stability_individual}

We next ask whether the disagreement that remains is structured, by testing whether a rater's departure from consensus replicates robustly \emph{within} that rater. For each rating, we first compute the deviation between a participant's standardized profile and the mean profile of the other participants who rated the same scenario. We then correlate a participant's mean deviation profile across two disjoint halves of the scenarios that they rated, averaging over 20 random splits. The two halves agree at $r=0.659$ (Pearson correlation), against a null of $r=-0.016$ obtained by permuting rater
identity ($n=200$, $p=0.005$), showing significant above chance stability of individual rating structure. 

We also test whether this agreement is due in part to idiosyncrasies of scale usage for a given participant. We find that this effect is unchanged after removing each participant's own mean ($r=0.659$) and their mean and standard deviation ($r=0.691$), eliminating the possibility that stability across the two split halves is due only to \textit{how} the participant uses the scale. Thus, the variability observed between raters is largely structured rather than noise, with individual differences in how scenarios are appraised replicated across two random and disjoint halves of the scenarios a single participant rates. 

\subsection{Agreement with Theory}
\label{sec:app_agreement_theory}

\begin{table*}[t]
\centering
\caption{Directional alignment of the human \emph{mean} appraisal profile with meta-analytic theory: Pearson correlation between each emotion's mean deviation profile and the theory's signed-$r$ vector. Higher is better; $^{\dagger}$ marks $p>0.05$ (positive but non-significant).}
\label{tab:theory_profile}
\setlength{\tabcolsep}{4pt}
\resizebox{\textwidth}{!}{%
\begin{tabular}{lcccccccccccc}
\toprule
Emotion & hope & happiness & disgust & anger & interest & fear & sadness & challenge & guilt$^{\dagger}$ & surprise$^{\dagger}$ & pride$^{\dagger}$ & boredom$^{\dagger}$ \\
\midrule
Profile $r$ & 0.997 & 0.989 & 0.953 & 0.944 & 0.920 & 0.888 & 0.769 & 0.758 & 0.534 & 0.476 & 0.465 & 0.304 \\
\bottomrule
\end{tabular}%
}
\end{table*}

\begin{table*}[t]
\centering
\caption{Per-scenario directional sign agreement with theory: percentage of mapped appraisal dimensions on which individual human scenario ratings fall on the theory-predicted side of the grand mean. Chance is 50\%.}
\label{tab:theory_scenario}
\setlength{\tabcolsep}{4pt}
\resizebox{\textwidth}{!}{%
\begin{tabular}{lcccccccccccccc}
\toprule
Emotion & shame & happiness & contempt & hope & interest & guilt & pride & fear & boredom & disgust & anger & sadness & surprise & challenge \\
\midrule
Agreement (\%) & 100.0 & 91.7 & 88.2 & 81.9 & 78.9 & 73.0 & 71.7 & 71.1 & 69.1 & 69.1 & 63.6 & 63.2 & 60.8 & 47.6 \\
\bottomrule
\end{tabular}%
}
\end{table*}

Beyond testing the reliability of ratings in an unsupervised manner, we investigate whether the obtained human data aligns with theoretical expectations from prior research in human psychology. We test whether human appraisals align with the directional predictions of an existing, large-scale meta-analytic study~\cite{yeo2024associations} that specifies theoretically expected associations between emotions and appraisal dimensions. Specifically, as described in the main body, the study provides for a given emotion category $e$ and appraisal dimension $d$, a signed correlation $r_{e,d}$ indicating whether $d$ should be high ($r_{e,d}>0$) or low ($r_{e,d}<0$) for that emotion. We retain only predictions that assert a direction, $|r_{e,d}| \geq 0.10$, mapping the dimensions from the theory onto our appraisal dimensions, averaging where a dimension maps to multiple reverse-coded dimensions such as \emph{legitimacy-cheated} and \emph{legitimacy-fair}. The results from the study also include emotion categories overlapping with 12 out of the 15 emotions used in our set, which we include for this analysis. We use each dimension's mean rating (across emotions) $\bar{g}_d$ as the reference point for determining the ``high'' or ``low'' direction.

We first ask whether an emotion's average appraisal profile leans the way theory predicts: for each emotion we correlate its human mean deviation profile $\mathbf{x}_e = \boldsymbol{\mu}_e - \bar{\mathbf{g}}$ with the theory vector $\mathbf{r}_e$ across shared dimensions. Human means are strongly aligned with theory for the majority of emotions (Table~\ref{tab:theory_profile})---the profile correlation exceeds $0.75$ for eight of twelve emotions (all significant at $p<0.05$). The remaining emotions (guilt, surprise, pride, boredom) show positive but non-significant correlations. 

Next, we ask whether \emph{individual scenarios}, not just the mean, replicate the theoretically predicted associations. For each scenario $s$ of emotion $e$ we score the fraction of dimensions on which the human rating falls on the predicted side of the grand mean,
\[
A_s = \frac{1}{|D_e|}\sum_{d \in D_e} \mathbb{1}\!\left[\big(x_{s,d} > \bar{g}_d\big) = \big(r_{e,d} > 0\big)\right],
\]
and average $A_s$ within each emotion. Sign agreement exceeds the $50\%$ chance level for every emotion but challenge, where the agreement is $47.6\%$(Table~\ref{tab:theory_scenario}). This establishes that human appraisals are largely directionally consistent with theory both on average and scenario-by-scenario, supporting the interpretation that their scenario-to-scenario variability reflects valid signal rather than noise.

\subsection{Aggregate Emotion Profile Reliability}
\label{sec:app_agg_emotion_profile}

Finally, we verify reliability of the human appraisal data at the level of aggregated profiles for each emotion category, as our analyses relies primarily on emotion-level appraisal distributions rather than on individual ratings. Partitioning participants into disjoint halves and comparing the resulting per-emotion mean profiles over 500 random partitions yields a split-half correlation of $r=0.911$ ($95\%$ CI  $[0.877, 0.934]$). The emotion representations on which our comparisons rest are thus highly reproducible, even though individual raters vary in a stable manner around individual scenarios for each emotion category. 

\section{Eliciting LLM Responses}
\label{sec:app_llm_data}

We include 8 models for our main analyses. For all proprietary models used, and DeepSeek, we access them through their respective API providers: for GPT-based models, we use \texttt{https://openai.com/api/}, for Gemini-based models, we use \texttt{https://aistudio.google.com/}, and for DeepSeek, we use \texttt{https://platform.deepseek.com/}. For Muse Glimmer, we use the Groq API (\texttt{https://console.groq.com/}). All of the other models are inferenced locally, using HuggingFace (\texttt{https://huggingface.co/}) and the \texttt{vllm} library. The models are hosted using a single or 2 (for QwQ 32B) A100 GPUs, on a shared Slurm cluster. The total inference time for a single set of inferences over all 4590 (= 270 scenarios $\times$ 17 appraisal questions) is between 1 to 5 hours for different models, with reasoning models taking longer. Model responses are parsed using a curated parser code, with parsing rates above 98\% for all models under all conditions. Responses that fail to parse are excluded from the analysis. 

\section{Coarse Similarity in Structure with PCA}
\label{sec:app_pca}

\begin{wrapfigure}{r}{0.60\textwidth}
    \centering
    \begin{subfigure}{0.29\textwidth}
        \includegraphics[width=\linewidth]{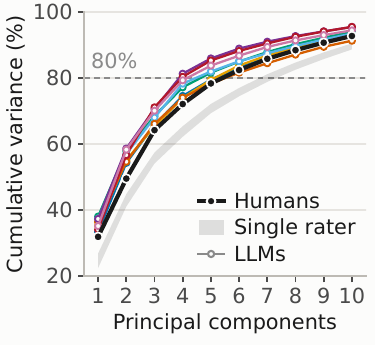}
        \caption{}
        \label{fig:pca_cumulative_variance}
    \end{subfigure}
    \begin{subfigure}{0.29\textwidth}
        \includegraphics[width=\linewidth]{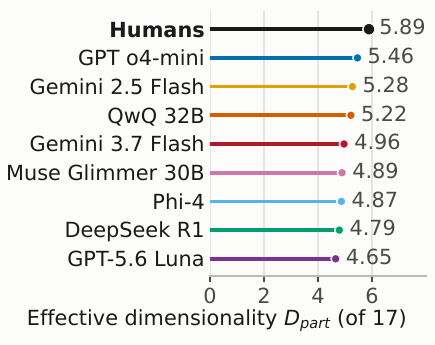}
        \caption{}
        \label{fig:pca_eff_dims}
    \end{subfigure}
    \caption{Coarse correspondence in appraisal-based representation space for LLMs and humans. (a) shows the cumulative variance explained as a function of the number of principal components, (b) shows the effective number of dimensions used. For Fig. (a), beyond showing the variance explained from averaged human ratings, we also show an approximation of the variance if each scenario was rated by a single rater (gray band). For this, for each scenario, we sample a single rater at random to compute PCA and repeat the process 100 times to show the 95\% interval.}
    \label{fig:pca_full}
\end{wrapfigure}

\textbf{LLMs and humans show coarse similarity in effective dimensions representing emotion concepts.} We first use Principal Component Analysis (PCA), establishing whether models and humans use a similar low-dimensional structure of appraisals to represent emotions. We find that for both humans and models, a small number of components captures $>50\%$ of the variance(Fig.~\ref{fig:pca_cumulative_variance}). Within this shared low-dimensional structure, however, consistently higher variance is captured, with a given number of principal components, for LLMs, compared to humans. All models further use a marginally lower number of effective dimensions compared to humans (Fig.~\ref{fig:pca_eff_dims}), while humans concur with theoretical expectations~\citep{smith1985patterns}, using $\approx6$ effective dimensions. Thus, although models and humans share a coarsely similar low-dimensional structure (i.e., using a similar number of effective dimensions), they use a more compressed representation space.

\begin{figure}
    \centering
    \includegraphics[width=0.9\linewidth]{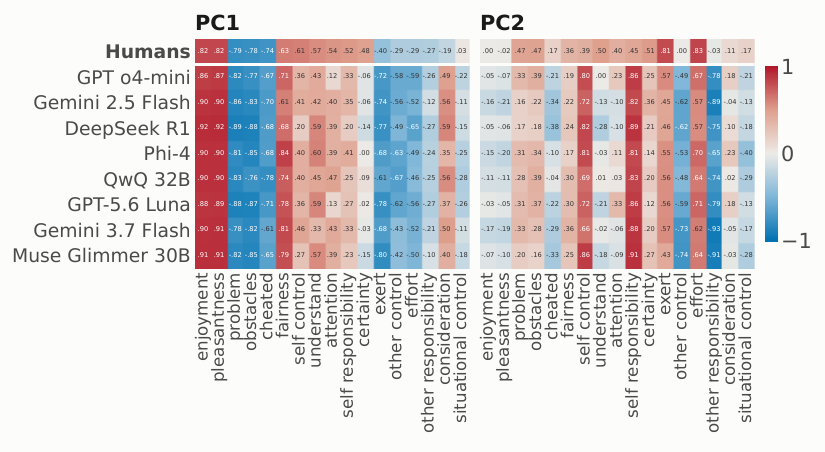}
    \caption{The complete loadings of each original appraisal dimension ($n=17$) onto the top two principal components recovered (left = PC1, right = PC2).}
    \label{fig:pca_1_2_full}
\end{figure}

In this section, in addition to the results about low-dimensional structure of representations in models and humans, we examine qualitative differences in the appraisal distributions. Studying the correlation of each original appraisal dimension with the top two principal components (PCs) recovered, we note a key similarity, and a difference: in both humans and LLMs, the first PC is driven by valence-related dimensions, such as pleasantness and enjoyment. However, models and humans differ in which appraisal dimension correlates the most strongly with the second PC. In humans, anticipated effort (congruent with the common ``arousal" dimension~\citep{smith1985patterns}) drives the second PC, while for models, agency-related dimensions such as self- or other-responsibility load most strongly. We also show the complete loadings in Fig.~\ref{fig:pca_1_2_full}.

\section{Diversity and Discreteness using Wasserstein Distance}
\label{sec:app_separation_ratio}

Here, we first provide the formal definitions of the distance measures we compute to quantify representation spread within a given emotion category ($\mathrm{D_{Within}}$) and distance between the representations from two distinct emotion categories ($\mathrm{D_{Between}}$). Formally, 
\begin{equation}
\mathrm{D_{Between}} = \frac{1}{\binom{|\mathcal{E}|}{2}} \sum_{\substack{e,f \in \mathcal{E} \\ e < f}} W_2\!\left(X_e, X_f\right),
\end{equation}
\begin{equation}
\mathrm{D_{Within}} = \frac{1}{|\mathcal{E}|}\sum_{e\in\mathcal{E}} \frac{1}{S}\sum_{s=1}^{S} W_2\!\left(X_e^{(s,1)},\, X_e^{(s,2)}\right),
\end{equation}
\begin{equation}
\mathrm{Separation} = \frac{\mathrm{D_{Between}}}{\mathrm{D_{Within}}},
\end{equation}
where $\mathcal{E}$ is the set of emotions, $X_e$ the set of standardized scenario-level appraisal profiles labeled with emotion $e$, and $X_e^{(s,1)}, X_e^{(s,2)}$ the two halves of $X_e$ under random split $s$.

\paragraph{Controlling for number of random halves. } In computing the main results, we set the value of the parameter $s = 30$. In Table~\ref{tab:app_separation_ratio_split_sensitivity}, we show the effect of varying this parameter. Specifically, we note the effect of computing the separation ratio by setting $s=50, 70, 100$ times. We find the separation ratio is insensitive to the number of random half-splits used to estimate the within-emotion spread. The largest change between $s=30$ and $s=100$ is an increase in the ratio of $0.021$ for Phi-4, indicating that increasing the number of random halves considered, if at all, decreases the within-emotion diversity.

\begin{table}[t]
\centering
\small
\begin{tabular}{lcccc}
\toprule
 & \multicolumn{4}{c}{Separation Ratio}  \\
\cmidrule(lr){2-5}
& $s=30$ & $s=50$ & $s=70$ & $s=100$ \\
\midrule
\textbf{Humans} & 1.293 & 1.298 & 1.296 & 1.298 \\
\midrule
GPT o4-mini & 2.084 & 2.108 & 2.083 & 2.092 \\
Gemini 2.5 Flash & 2.196 & 2.198 & 2.190 & 2.193 \\
DeepSeek R1 & 2.275 & 2.282 & 2.269 & 2.266 \\
Phi-4 & 2.130 & 2.160 & 2.133 & 2.151 \\
QwQ 32B & 1.896 & 1.906 & 1.898 & 1.900 \\
GPT-5.6 Luna & 2.429 & 2.450 & 2.431 & 2.441 \\
Gemini 3.7 Flash & 2.529 & 2.549 & 2.540 & 2.524 \\
Muse Glimmer 30B & 2.402 & 2.433 & 2.412 & 2.413 \\
\bottomrule
\end{tabular}
\caption{Robustness of the separation ratio with different random seeds, and different numbers of random halves used to estimate the within-emotion spread. Separation ratio is recomputed at $s=30,50,70,100$.}
\label{tab:app_separation_ratio_split_sensitivity}
\end{table}

\paragraph{Controlling for Rater-reliability.}
Each human appraisal representation for a given scenario is an average over the participants who rated that scenario, of whom there are $3.7$ on average and never more than six. Every individual rating carries some error particular to the rater, and averaging more participants cancels more of it, so the within-emotion spread we measure depends on how many people happened to rate each scenario. A model has no counterpart to this, since it produces one fixed rating per scenario and dimension. To establish how much of the human--model difference this asymmetry could explain, we recomputed the human profiles from only $k$ randomly chosen participants per rating, for $k$ from one to five, repeating each draw five times. Where a scenario had fewer than $k$ participants available we used all of them, so the average number of participants actually used is reported alongside $k$.  Let $W_{\mathrm{within}}(k)$ denote the within-emotion spread computed from $k$-participant averages, and let $p=17$ be the number of appraisal dimensions. If each rating is a true value plus independent rater error of variance $\sigma^2$, then averaging $k$ participants leaves the true value unchanged and reduces the error variance to $\sigma^2/k$ ~\citep{spearman1910correlation,brown1910some,shrout1979intraclass}. The scenarios belonging to one emotion then form a cloud whose covariance is $\Sigma + (\sigma^2/k)\mathbf{I}$, where $\Sigma$ is the covariance that would be observed without rater error, so the total variance of the cloud is inflated by $p\sigma^2/k$. The squared Wasserstein spread is, to a good approximation, proportional to that total variance, which provides 
\begin{equation}
  W_{\mathrm{within}}^{2}(k) \;=\; A_{\mathrm{within}} + \frac{B_{\mathrm{within}}}{k},
  \label{eq:sb}
\end{equation}
with $A_{\mathrm{within}}$ the squared spread of the noise-free cloud and $B_{\mathrm{within}}$ absorbing the error variance. We estimate the two coefficients by ordinary least squares across the five sampled depths and report $\sqrt{A_{\mathrm{within}}}$, the spread expected of a panel of unlimited size. Note that the error term is additive in variance rather than in distance, which is why Eq.~\ref{eq:sb} is written for the squared spread. Applying the same fit to the between-emotion distance yields $A_{\mathrm{between}}$, and $\sqrt{A_{\mathrm{between}}/A_{\mathrm{within}}}$ is the separation ratio at the same limit. Setting $1/k=0$ extrapolates beyond every panel size we observe, so this estimate relies on the $1/k$ form of Eq.~\ref{eq:sb} continuing to hold.

Averaging more participants makes the human profiles more precise and reduces their within-emotion spread, from $4.36$ when each scenario is rated by a single participant to $3.60$ at the full number of raters available. The distance between emotions
is almost unaffected. The human separation ratio therefore rises with rater panel size, from $1.089$ to $1.293$, and extrapolating to a panel of unlimited size
gives $1.387$ (Table~\ref{tab:rater-depth}). All eight models lie above that value, ranging from $1.896$ for QwQ 32B to $2.529$ for Gemini 3.7 Flash.
 
We also repeat the comparison of each model against humans separately at each panel size, using Wilcoxon's signed-rank test for statistical significance, and reporting rank-biserial correlation as effect size. Every model differs significantly from humans on both quantities, at every panel size. For the distance between emotions all corrected $p$-values are below $3\times10^{-7}$, with effect sizes of at least $0.58$. For the within-emotion spread every model is more compressed than humans on all fifteen emotions, with an effect size of $-1.00$ and a corrected $p$-value of
$4.9\times10^{-4}$. The difference between humans and models therefore does not depend on how many people rated each scenario. It holds even when each human profile is built from a single participant, which is the least precise estimate the data allow and the one most favorable to the models.

\begin{table}[t]
\centering
\small
\resizebox{\textwidth}{!}{
\begin{tabular}{lcccc}
\toprule
raters drawn & mean depth & between-emotion & within-emotion & separation \\
per cell, $k$ & achieved & $W_2$ & $W_2$ & ratio \\
\midrule
1 & 1.00 & 4.745 & 4.356 & 1.089 {\tiny$\pm$0.007} \\
2 & 1.97 & 4.736 & 4.014 & 1.180 {\tiny$\pm$0.009} \\
3 & 2.83 & 4.674 & 3.762 & 1.243 {\tiny$\pm$0.004} \\
4 & 3.36 & 4.661 & 3.646 & 1.279 {\tiny$\pm$0.005} \\
5 & 3.66 & 4.654 & 3.600 & 1.293 {\tiny$\pm$0.001} \\
\midrule
all available & 3.67 & 4.653 & 3.598 & \textbf{1.293} \\
$k\rightarrow\infty$ (extrapolated) & --- & 4.634 & 3.341 & \textbf{1.387} \\
\midrule
\multicolumn{4}{l}{\emph{range of models, for reference}} & 1.896 -- 2.529 \\
\bottomrule
\end{tabular}}
\caption{\textbf{The human--model gap in the separation ratio is not an artefact of rater noise.} Human appraisals were recomputed from $k$ raters drawn at random per scenario and dimension, five draws at each depth; where a cell held fewer than $k$ raters all of them were used, so the depth actually achieved is reported beside the requested one. Averaging more raters removes measurement error and compresses the within-emotion spread, raising the ratio from $1.089$ at a single rater to $1.293$ at the full panel. Extrapolating to a perfectly reliable panel by fitting $W_2^{2}(k)=A+B/k$ (within-emotion: $A=11.17$, $B=8.09$, $R^{2}=0.95$; between-emotion: $A=21.47$, $B=1.15$, $R^{2}=0.71$) yields $1.387$, still well below the least discrete model. Values after $\pm$ are standard deviations across the five draws. }
\label{tab:rater-depth}
\end{table}

\paragraph{Introducing sampling noise in models. }

The human representation for each scenario is the average of about 3.7
ratings, whereas each model profile is a single response.  Because human participants might disagree with each other, the average scenario representation could change if a different set of participants had rated the scenario. We call this variation the \textit{measurement error} of the human profile. For a single appraisal dimension on one scenario (say, a single cell $c$), let $s_c$ be the standard deviation of the participants' ratings for that cell and $n_c$ the number of participants who rated it. Participants were assigned to scenarios at random and rated them independently, so the average of $n_c$ ratings varies from one set of raters to another with standard deviation $s_c/\sqrt{n_c}$, the standard error of
the mean. This is the size of the measurement error in cell $c$: how far the average of the participants we observed is likely to be from the average over all possible raters. A model profile is a single response and is treated as exact.

Measurement error is unrelated to the emotion category, so it makes the representations of scenarios from the same emotion differ more from one another, which could potentially enlarge the within-emotion spread. Because normalizing ratings prior to any analyses fixes the total variance of each dimension, it also reduces the between-emotion distance. Part of the difference between humans and models could therefore come from this error rather than indicating genuine difference in how concepts are represented. To remove it and enable a stricter comparison, we add Gaussian noise with standard deviation $s_c/\sqrt{n_c}$ to each model rating, so that model profiles carry the same measurement error as human profiles. For cells
rated by a single participant, where $s_c$ is undefined, we use the mean standard deviation of that dimension. We then repeat the full analysis unchanged: standardization of ratings, followed by computing the 105 between-emotion and 15 within-emotion $W_2$ distances, and the two Wilcoxon signed-rank tests against humans with Holm correction across the eight models. The human data are left exactly as in the main analysis. We repeat this for five independent noise draws and report the mean.

Table~\ref{tab:noise_separation} reports the result.
Although matching the measurement error moves every model relatively towards humans, between-emotion distance of every model remains significantly larger than the human one in all five draws (Holm-corrected $p < 0.001$; rank-biserial correlation 0.81--0.93), and the within-emotion spread remains significantly smaller than the human one in all five draws for seven of the eight models too
(largest Holm-corrected $p = 0.011$). For QwQ 32B it is significant in three of the five draws (Holm-corrected $p$ from 0.008 to 0.12), although its spread is still smaller than the human spread for most emotions in every draw (rank-biserial correlation from $-0.47$ to $-0.75$). Measurement error therefore cannot explain the difference in representation structures observed between humans and models: with matched error, every model still places emotions further apart than humans do, and for every model except QwQ 32B the smaller within-emotion spread also remains significant.

\begin{table}[t]
\centering
\small
\setlength{\tabcolsep}{4pt}
\resizebox{\columnwidth}{!}{
\begin{tabular}{@{}lccccccc@{}}
\toprule
 & \multicolumn{2}{c}{Between-emotion} & \multicolumn{2}{c}{Within-emotion}
 & \multicolumn{2}{c}{Ratio} & Significant \\
\cmidrule(lr){2-3}\cmidrule(lr){4-5}\cmidrule(lr){6-7}
 & Default & Noise-matched & Default & Noise-matched & Default & Noise-matched & draws (B / W) \\
\midrule
Humans           & 4.65 & --   & 3.60 & --   & 1.29 & --    & --        \\
\midrule
Gemini 3.7 Flash & 5.44 & 5.31 & 2.15 & 2.96 & 2.53 & 1.80 & 5 / 5 \\
GPT-5.6 Luna     & 5.38 & 5.23 & 2.22 & 3.16 & 2.43 & 1.66 & 5 / 5 \\
Muse Glimmer 30B & 5.43 & 5.28 & 2.26 & 3.07 & 2.40 & 1.72 & 5 / 5 \\
DeepSeek R1      & 5.42 & 5.30 & 2.38 & 3.06 & 2.28 & 1.73 & 5 / 5 \\
Gemini 2.5 Flash & 5.39 & 5.29 & 2.45 & 3.08 & 2.20 & 1.72 & 5 / 5 \\
Phi-4            & 5.32 & 5.21 & 2.50 & 3.24 & 2.13 & 1.61 & 5 / 5 \\
GPT o4-mini      & 5.36 & 5.23 & 2.57 & 3.25 & 2.08 & 1.61 & 5 / 5 \\
QwQ 32B          & 5.31 & 5.17 & 2.80 & 3.43 & 1.90 & 1.51 & 5 / 3 \\
\bottomrule
\end{tabular}}
\caption{Separation of emotion concepts with the models measured as noisily as the human ratings, averaged across five noise draws. Between-emotion: mean $W_2$ distance between two emotion categories.
Within-emotion: mean $W_2$ spread within one category. Ratio: between divided by within. Significant draws: the number of the five draws in
which the model differs from humans at Holm-corrected $p<0.05$ (Wilcoxon signed-rank test) on the between-emotion distances (B) and on the within-emotion spreads (W).}
\label{tab:noise_separation}
\end{table}

\begin{figure}
    \centering
    \includegraphics[width=0.8\linewidth]{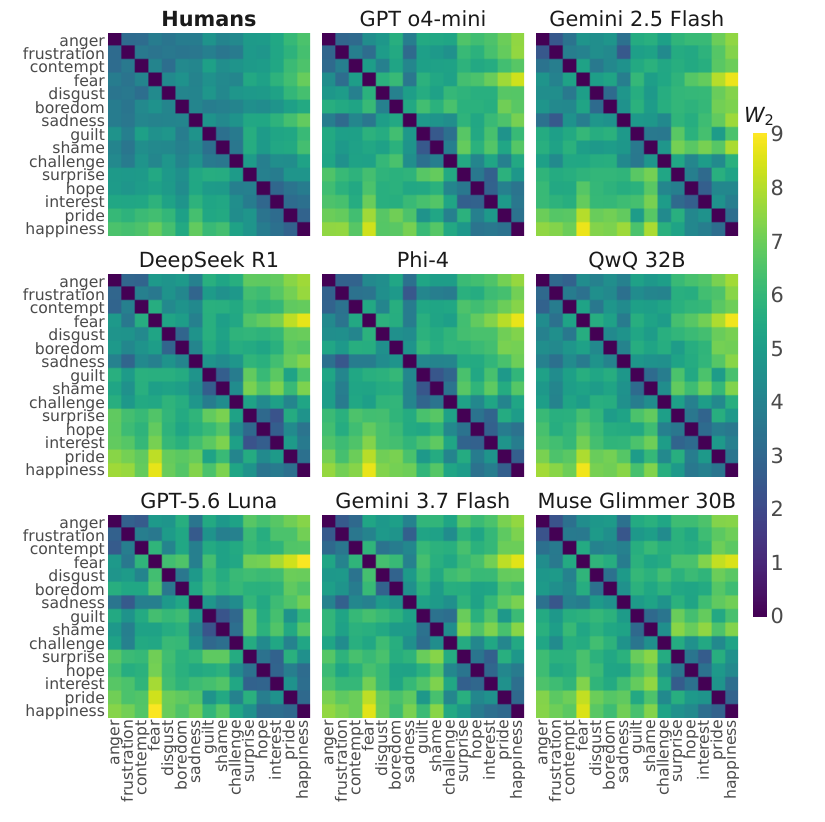}
    \caption{Visualization of multivariate $W_2$ distances between representations of different emotion concepts. In humans, lesser clear categories are observable, with lower distances between different emotions. On the other hand, for all LLMs, clearer, discrete clusters of emotions are observed.}
    \label{fig:wasserstein_heatmap_appendix}
\end{figure}

\paragraph{Visualizing Wasserstein distances.} In addition to the formulation of the distances and the comparisons based on statistical tests, we visualize the Wasserstein distances between different emotion categories, for each agent in Fig.~\ref{fig:wasserstein_heatmap_appendix}. In humans, the relative separation between different emotion concepts can be observed to be smoother and graded. For all LLMs, and in particular the frontier models, clearer clusters can be observed. For example, negative other-oriented emotions such as anger, frustration, or contempt are separated strongly from self-oriented negative emotions such as guilt or shame. 

We further separate the 15 emotion categories into different theoretically-aligned groups~\citep{smith1985patterns}, along the axes of different appraisal dimensions, to understand whether different groups of emotions are separated more or less from each other. Specifically, we group emotions along the following axes: valence (positive or pleasant vs. negative or unpleasant), agency (who is responsible for the situation: self, another person, or circumstances), anticipated effort (low effort vs. high effort), control/coping potential (controlled by self vs. situation), certainty (low vs. high) and fairness (fair vs. unfair). 

We find that both humans and models separate emotion categories strongly and similarly along the axes of valence, control, and fairness. We note, however, that separation along valence and fairness leads to highly overlapping groups of emotions: all positive emotions are considered fair, while most negative emotions are considered unfair. Cross-valence peairs are about $1.45$ times further apart than same-valence pairs for humans, and $1.39$ to $1.51$ times for the models. Further, both LLMs and humans do not separate emotions strongly based on anticipated effort. Humans do not separate emotions based on certainty at all (the average between-emotion distance across low and high-certainty groups are identical), while models show slightly stronger separation along certainty. The difference between models and humans, however, is the strongest for separation based on agency-related dimensions. For humans, agency barely separates emotions: cross-agency pairs are only 1.06 times further apart than same-agency pairs, which is not significant ($p=0.39$). For every model, the ratio is higher, 1.11 to 1.19. Thus, the significantly larger between-emotion distances suggest every model stretches cross-agency pairs more than same-agency pairs.

\section{Distinguishability of Emotion Concepts Using Behavioral Representations}
\label{sec:app_f1}

\paragraph{Implementation details of Logistic Regression.} To examine how distinguishable different emotion concepts are based on their behavioral representations, we simply fit for every agent an $\ell_2$-regularized logistic classifier for our main results. This setup allows predicting a scenario's emotion category from its 17-dimensional appraisal representation. We estimate determinacy of emotions through F1 scores achieved by the classifier, across three folds of stratified cross-validation, in a one-versus-all setup. We use $\ell_2$ regularization with the default strength of $C=1$, and no class weighting, given the comparable number of samples from each class. Implementation is done using the \texttt{scikit-learn} library. 

\begin{figure}
    \centering
    \includegraphics[width=0.8\linewidth]{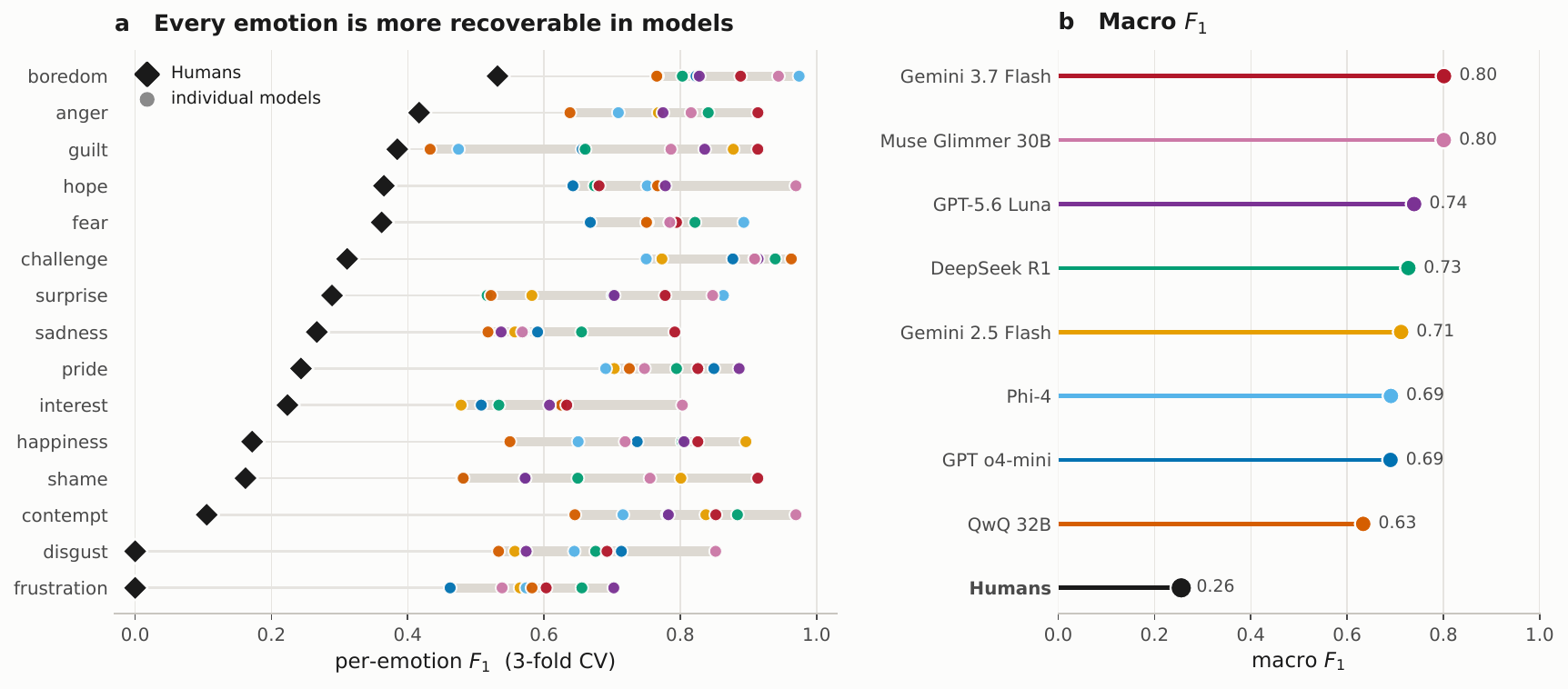}
    \caption{Complete F1 results. (a) Per-emotion F1 scores across different entities, where for each of the 15 emotion categories, F1 scores are higher for each model compared to humans. (b) Macro-F1 scores shown for each entity.}
    \label{fig:app_f1_full}
\end{figure}

\paragraph{Full prediction results.} We show the full results in Fig.~\ref{fig:app_f1_full}. Each emotion category is more easily separable and distinguishable from LLMs' behavioral representations than from human representations. Corroborating our previous results, representations from the newest frontier models (Muse Glimmer, Gemini 3.7 Flash and GPT 5.6 Luna) lead to the highest separation of emotion concepts (macro-F1 $\in [0.74, 0.80]$). 

Humans and LLMs also agree weakly on which emotion concepts are the most separable ($r=0.53, p < 0.05$). Across all LLMs, the most determinate emotions are challenge, boredom and contempt, while the least separable ones include frustration, sadness and interest. For humans, on the other hand, the most determinate are boredom, anger and guilt, and the least separable are frustration and disgust. For humans, half of all disgust scenarios are recovered as contempt and $47\%$ of shame scenarios as guilt, with representations of interest and surprise, and anger and frustration, overlapping similarly. LLM representations of the same pairs also overlap but much less often, with the logistic classifier labeling disgust as contempt $17\%$ of the time and shame as guilt $23\%$ of the time.

\begin{table}[t]
\centering
\small
\setlength{\tabcolsep}{5pt}
\begin{tabular}{@{}lccccc@{}}
\toprule
 & \multicolumn{2}{c}{Macro $F_1$} & \multicolumn{2}{c}{Difference from humans} & Significant \\
\cmidrule(lr){2-3}\cmidrule(lr){4-5}
 & Default & Noise-matched & Default & Noise-matched & draws \\
\midrule
Humans           & 0.26 & --   & --   & --   & -- \\
\midrule
Gemini 3.7 Flash & 0.80 & 0.69 & 0.55 & 0.43 & 5 \\
Muse Glimmer 30B & 0.80 & 0.66 & 0.55 & 0.40 & 5 \\
GPT-5.6 Luna     & 0.74 & 0.57 & 0.48 & 0.32 & 5 \\
DeepSeek R1      & 0.73 & 0.62 & 0.47 & 0.36 & 5 \\
Gemini 2.5 Flash & 0.71 & 0.62 & 0.46 & 0.37 & 5 \\
GPT o4-mini      & 0.69 & 0.57 & 0.44 & 0.32 & 5 \\
Phi-4            & 0.69 & 0.54 & 0.44 & 0.28 & 5 \\
QwQ 32B          & 0.63 & 0.52 & 0.38 & 0.27 & 5 \\
\bottomrule
\end{tabular}
\caption{Recovery of the emotion label from the appraisal ratings when the same measurement error from humans is added to the model ratings. Macro $F_1$: the average of the fifteen per-emotion $F_1$ scores of the cross-validated classifier, chance is $0.07$. Significant draws: the number of the five draws in which the model's per-emotion scores are higher than the human ones at Holm-corrected $p<0.05$ (Wilcoxon signed-rank test over the fifteen emotion categories).}
\label{tab:noise_f1}
\end{table}

\paragraph{Introducing sampling noise in models.} As in Appendix~\ref{sec:app_separation_ratio}, we add Gaussian noise to the model ratings so that model representations carry the same \textit{measurement error} as human profiles, here $s_c/\sqrt{n_c}$ per scenario and appraisal dimension. We then repeat the same analysis as in the main results: standardization, the same
logistic regression with stratified three-fold cross-validation, the per-emotion $F_1$ scores and their macro average. The human data are left unchanged as in the main analysis, and we report the mean of five independent noise draws.

Matching the human-level \textit{measurement error} lowers every model's macro $F_1$ to some extent, from $0.63$--$0.80$ to $0.52$--$0.69$, against $0.26$ for humans (Table~\ref{tab:noise_f1}). However, $65$--$80\%$ of each model's gap to humans remains, and $73\%$ of the mean gap. Every model's per-emotion scores remain significantly higher (paired Wilcoxon signed-rank test over the fifteen emotions with Holm correction) than the human ones in all five draws, and every one of the fifteen emotions is still recovered better in every model than in humans.

\section{Consistency across implicit and explicit judgments}
\label{sec:app_implicit_explicit}

\paragraph{Details of the Setup.} In addition to the default setup where models and humans provide appraisal ratings (along 17 dimensions) for each scenario in our benchmark dataset, we evaluate a distinct elicitation format involving explicit judgments about what appraisal dimension separates a given pair of emotions. Here, we describe the data collection process from both humans and models. Prior to data collection, in consultation with psychology experts, we subsample 90 emotion pairs out of the possible 105 pairs in our dataset ($15\choose2$), excluding emotion pairs that are highly similar (e.g., guilt and shame) and pairs where a single appraisal dimension alone is the strong separator (e.g., valence for happiness and anger).  

Models are required to rate the importance of each appraisal dimension in separating a given pair of emotions. For each pair of emotions, 6 responses are sampled from each model: the first three for general consistency of ranking, and the second three testing the same pair of emotions but presenting the emotion categories in the prompt in the reversed order. For example, for the emotions \textit{happiness} and \textit{pride}, the first three prompts ask models to rank how appraisals distinguish ``\textit{happiness and pride}'', while the other three prompts frame it as ``\textit{pride and happiness}''. This leads to a total of 90 $\times$ 6 = 540 responses sampled from each model. We use the average ranking of these 6 independently sampled responses for our analyses. The rating scale for each appraisal dimension ranges from 1--5, given that we expect only a few appraisal dimensions to be the most relevant for distinguishing between any given pair of emotions. 

For humans, explicit judgments of which appraisal dimensions separate a given pair of emotions are also collected similarly. The questions about pairwise discrimination are provided to human participants after they provide ratings to the appraisal questions. Each participant rates 10 pairs, drawn randomly from the possible pair of emotion pairs (90). We apply the same data filtration process as for the main data collected, and retain $\approx5$ human ratings per emotion pair. The average of these ratings across the participants who rated a pair constitutes the human explicit judgment for that pair. The implicit judgment is computed from the individual participant ratings rather than from scenario averages, because the implicit separation between two emotions on a dimension is measured as the difference between the distribution of ratings given to one emotion and the distribution given to the other, which requires the individual responses; averaging within a scenario first would alter the spread of these distributions. The two sides of the comparison are therefore aggregated differently: the explicit judgment is a mean over participants, while the implicit judgment retains each participant's rating.

\paragraph{Method to compute correlations.} Formally, an agent rates, for every emotion pair $(A,B)$ and dimension $d$, how strongly $d$ differentiates $A$ from $B$ on a $1$--$5$ scale, yielding an explicit vector $\mathbf{e}_{AB}\in\mathbb{R}^{17}$. For each pair $(A,B)$ and dimension $d$ we already have the appraisal ratings along $d$ to $A$'s scenarios and to $B$'s scenarios, forming two groups $X^{d}_{A}$ and $X^{d}_{B}$. We then score how strongly $d$ separates $A$ and $B$, with the (absolute) Cliff's $\delta$,
\begin{equation}
  \bigl|\delta^{d}_{AB}\bigr|
  = \bigl|\Pr(X^{d}_{A} > X^{d}_{B}) - \Pr(X^{d}_{A} < X^{d}_{B})\bigr|
  \in [0,1],
\end{equation}
A value of $0$ means the two emotions are inseparable using $d$, while $1$ would indicate $d$ separates the emotions perfectly. Collecting these values across dimensions yields an implicit vector $\mathbf{i}_{AB}\in\mathbb{R}^{17}$ describing what actually distinguishes $A$ from $B$ in the agent's ratings. For each pair of emotions, we then rank the $17$ dimensions by $\mathbf{e}_{AB}$ and by $\mathbf{i}_{AB}$ and measure agreement with Spearman's rank correlation $\rho\bigl(\mathbf{e}_{AB},\mathbf{i}_{AB}\bigr)$, with a high $\rho$ indicating that the agent's stated discriminators match its behavioral ones.

\paragraph{Full results.} We show full results in Table~\ref{tab:implicit-explicit}. Humans show no correspondence between their explicit judgments of which appraisal dimension separates two emotions, and the same separation observed from their implicit behavioral representations. Models, on the other hand, show moderate-to-strong correspondence. Notably, the highest correspondence is observed for models within the Gemini family, followed by other frontier-capability models. QwQ 32B and Phi-4 show the least correspondence. 

\begin{table}[t]
\centering
\small
\setlength{\tabcolsep}{4.5pt}
\begin{tabular}{lccccccccc}
\toprule
 & \rotatebox{90}{\textbf{Humans}} & \rotatebox{90}{GPT o4-mini} & \rotatebox{90}{Gemini 2.5 Flash} & \rotatebox{90}{DeepSeek R1} & \rotatebox{90}{Phi-4} & \rotatebox{90}{QwQ 32B} & \rotatebox{90}{GPT-5.6 Luna} & \rotatebox{90}{Gemini 3.7 Flash} & \rotatebox{90}{Muse Glimmer 30B} \\
\midrule
mean $\rho$ & \textbf{0.058} & 0.401 & 0.508 & 0.496 & 0.289 & 0.240 & 0.356 & 0.554 & 0.423 \\
median $\rho$ & \textbf{0.077} & 0.425 & 0.508 & 0.490 & 0.330 & 0.293 & 0.396 & 0.548 & 0.445 \\
\bottomrule
\end{tabular}
\caption{Consistency of emotion representations across two different elicitation formats. The table reports the Spearman correlation between the two rankings, averaged over the $90$ pairs. Every model is above zero (Holm-corrected $p<<0.001$) and exceeds humans when paired on the same emotion pairs (all Holm-corrected $p<<0.001$; rank-biserial $0.55$ to $0.96$). The human value is not distinguishable from zero ($p=0.069$). The human correspondence computed separately for each individual participant, is $0.017$ on average ($p=0.61$).}
\label{tab:implicit-explicit}
\end{table}

\paragraph{Introducing sampling noise in models.} As in Appendix~\ref{sec:app_separation_ratio} and~\ref{sec:app_f1}, we add Gaussian noise to the model appraisal ratings to repeat our analyses under the same measurement error as the human ratings. The size of the measurement error depends on how many human ratings are averaged for a single sample (a given appraisal dimension for a given scenario). For the current analysis, as we mention above, we retain the pool of human ratings per scenario, instead of averaging. In the previous analyses (separation ratio, or recoverability of emotions using F1), the average human representation for a scenario is the mean of the $n_c$ participants who rated a scenario, so the error is $s_c/\sqrt{n_c}$. Here the implicit separation between two emotions on a dimension is the difference between the distribution of ratings given to one emotion and the distribution given to the other, so the individual participant ratings are pooled rather than first averaged within a scenario. The error for a single pooled rating thus encompasses the complete spread across raters, $s_c$. We consider the error for this pooled setup, repeating the analyses by correlating each agent's explicit ranking of the 17 dimensions for a pair with the implicit ranking recomputed from the noised ratings, and testing each model against
humans with a paired Wilcoxon signed-rank test over the emotion pairs with Holm correction across the eight models. We report the mean of five independent noise draws. 

The alignment between the explicit and the implicit ranking is almost unaffected by the added error (Table~\ref{tab:noise_ie}). With the injected error, $84$--$101\%$ of each
model's gap to humans remains. Every model
remains significantly above humans at both levels (Holm-corrected $p < 0.001$; rank-biserial correlation $0.55$--$0.99$). Three models show a
retained share slightly above $100\%$, meaning the noise improves their consistencty across response formats. 
 
\begin{table}[t]
\centering
\small
\setlength{\tabcolsep}{5pt}
\begin{tabular}{@{}lccc@{}}
\toprule
 & & \multicolumn{2}{c}{Noise-matched} \\
\cmidrule(lr){3-4}
 & Default & $\rho$ & Gap kept \\
\midrule
Humans           & 0.02 &--   & --    \\
\midrule
Gemini 3.7 Flash & 0.55 & 0.55 & 98\%  \\
Gemini 2.5 Flash & 0.51 & 0.44 & 86\%  \\
DeepSeek R1      & 0.50 & 0.48 & 96\%  \\
Muse Glimmer 30B & 0.42 & 0.42 & 101\% \\
GPT o4-mini      & 0.40 & 0.39 & 97\%  \\
GPT-5.6 Luna     & 0.36 & 0.32 & 89\%  \\
Phi-4            & 0.29 & 0.26 & 88\%  \\
QwQ 32B          & 0.24 & 0.21 & 84\%  \\
\bottomrule
\end{tabular}
\caption{Agreement between the explicit and the implicit ranking of the appraisal dimensions, with the model appraisal ratings given the same measurement error as the human ratings. $\rho$: the mean Spearman correlation over the emotion pairs between the two rankings. Gap kept: the share of the model's clean gap to the human value that remains under matched error. The human value differs here from Table~\ref{tab:implicit-explicit} as this is computed by pooling all ratings, instead of averaging.}
\label{tab:noise_ie}
\end{table}

\section{Determinacy using Interpretable discriminators}
\label{sec:app_synthtree}

\paragraph{Method.} In addition to the results of determinacy presented in the main body and in Appendix~\ref{sec:app_f1}, we use SynthTree~\citep{kuriabov2024synthtree} as an interpretable model framework, to probe hierarchical associations between appraisals and emotions, that makes different emotion concepts more or less distinguishable in different agents. SynthTree is a co-supervised teacher-student method: a black-box teacher supervises a set of sparse local models that are stitched together by a shallow decision tree, yielding a globally transparent predictor with a small accuracy loss relative to the teacher. We introduce the analyses based on SynthTree~\citep{kuriabov2024synthtree}, in addition to our main results, for the following reasons. Firstly, unlike logistic regression, which captures global average associations, SynthTree reveals \emph{hierarchical decision structure} and \emph{class-conditional feature usage} by combining tree-based partitioning with locally fitted linear models. Using SynthTree thus allows us to examine how dimensions are selectively emphasized in different regions of the appraisal space, and how this emphasis varies across models. Relatedly, it is also complementary to our main Wasserstein distance-based metrics~\ref{sec:results_alignment}, using a completely different model class (a tree of sparse local models), instead of a distance between distributions. Thirdly, we approach the analyses here using a complementary unit of analysis: a single human row is treated as the seventeen ratings \textit{one} participant provided to one scenario, with no pre-averaging over participants, followed by computing accuracy on held-out rows. This helps examine whether the main trends of determinacy robustly replicate when treating the human distribution in a complementary way. 

Similar to the original setting, we use the scenario-specific appraisal ratings as independent variables, with the emotion category as the dependent variable. Concretely, we partition the input (appraisal) space into cells by clustering. Within each cell, points are augmented and a sparse logistic model is fit to the teacher's predictions. We then grow and prune a decision tree over the cells so that each leaf carries one sparse local model. For every ordered pair of emotions $(A,B)$, we restrict an agent's ratings to the two emotions, label $B$ as the positive class, and train a SynthTree using the $17$ appraisal dimensions. To focus our comparison across agents on their appraisal-based representations rather than model selection, we fix the teacher to a random forest and use the same hyperparameters for all agents and pairs.

\paragraph{SynthTree Feature Importance (STFI).}
STFI quantifies how much each appraisal dimension contributes to an agent's fitted discriminator by functional perturbation of the fitted SynthTree~\citep{kuriabov2024synthtree}. Let $\hat f$ denote the fitted SynthTree's predicted probability of the positive emotion and $\{\mathbf{x}_i\}_{i=1}^{n}$ the ratings the tree was fit on. For dimension $k$ and a random permutation $\pi$ of $\{1,\dots,n\}$, let $\mathbf{x}_i^{(k,\pi)}$ replace the $k$-th coordinate of $\mathbf{x}_i$ with that of $\mathbf{x}_{\pi(i)}$. STFI is the mean squared change in prediction induced by this perturbation, averaged over $R$ permutations,
\begin{equation}
  \mathrm{STFI}(k)=\frac{1}{R}\sum_{r=1}^{R}\frac{1}{n}\sum_{i=1}^{n}\bigl(\hat f(\mathbf{x}_i)-\hat f(\mathbf{x}_i^{(k,\pi_r)})\bigr)^{2}.
\end{equation}
In a SynthTree, this admits a structural decomposition:
$\mathrm{STFI}(k)=\mathrm{FI}_{\mathrm{route}}(k)+\mathrm{FI}_{\mathrm{leaf}}(k)+\mathrm{FI}_{\mathrm{interact}}(k)$, which separates a dimension's effect through the tree's routing from its effect through the sparse leaf models---a decomposition unavailable for the black-box teacher. We use $R=20$ and normalize STFI to sum to one over the $17$ dimensions. Its \emph{concentration} is summarized by the Gini coefficient of the STFI vector (higher $=$ reliance on fewer dimensions). To relate what an agent \emph{uses} to what it \emph{does} and \emph{states} (similar to the comparison in Appendix~\ref{sec:app_implicit_explicit}), we compare the STFI profile against the implicit per-dimension discriminability (absolute Cliff's $\delta$ between the two emotions' ratings, $\mathbf{i}_{AB}$) and the agent's explicit stated discriminators ($\mathbf{e}_{AB}$), via Spearman rank correlation.

\begin{figure}[t]
  \centering
  \includegraphics[width=0.62\textwidth]{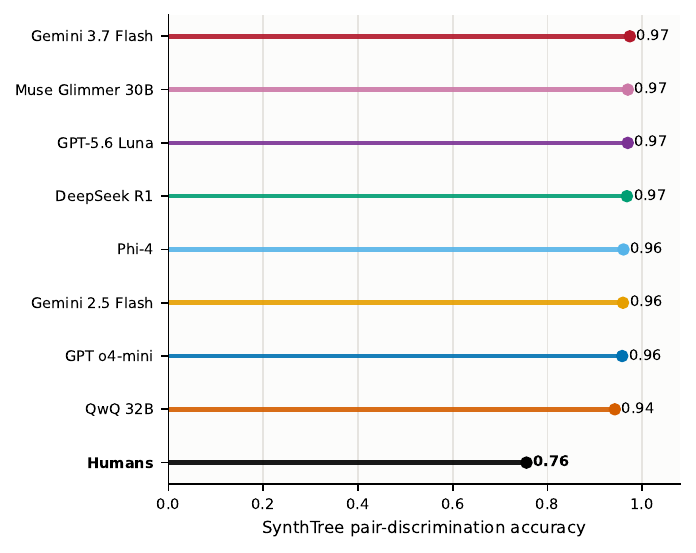}
  \caption{Mean balanced accuracy across all $105$ emotion pairs, providing a measure of how distinguishable the emotion concepts are in an agent's appraisal space.}
  \label{fig:separability}
\end{figure}

\paragraph{Emotion pairs are far easier to separate from LLMs' behavioral representations than from humans'.} Averaged across all possible 105 emotion pairs ($=$$15\choose{2}$), the cross-validated balanced accuracy of the fitted discriminator is $0.76$ for the human representations and between $0.94$ and $0.97$ for the eight models (Figure~\ref{fig:separability}, Table~\ref{tab:synthtree_summary}). Further, across \textit{all} possible emotion pairs and models, the separability (as measured by accuracy) with LLMs' behavioral representations exceeds that with humans, with the smallest margin $0.01$ on guilt against shame and the largest difference being $0.50$ on contempt against disgust. Across behavioral representations from all LLMs, perfect separability is achievable (balanced accuracy $=1.0$) on between $25$ to $52$ of the emotion pairs (out of $105$) whereas using the human representations, perfect separability is never achievable. The pairs that are hardest to separate based on human representations are contempt and disgust ($0.46$), anger and frustration ($0.51$) and frustration and sadness ($0.51$). All of these are pairs where both emotions share negative valence and other-orientation. 

\begin{figure}
  \centering
  \includegraphics[width=\textwidth]{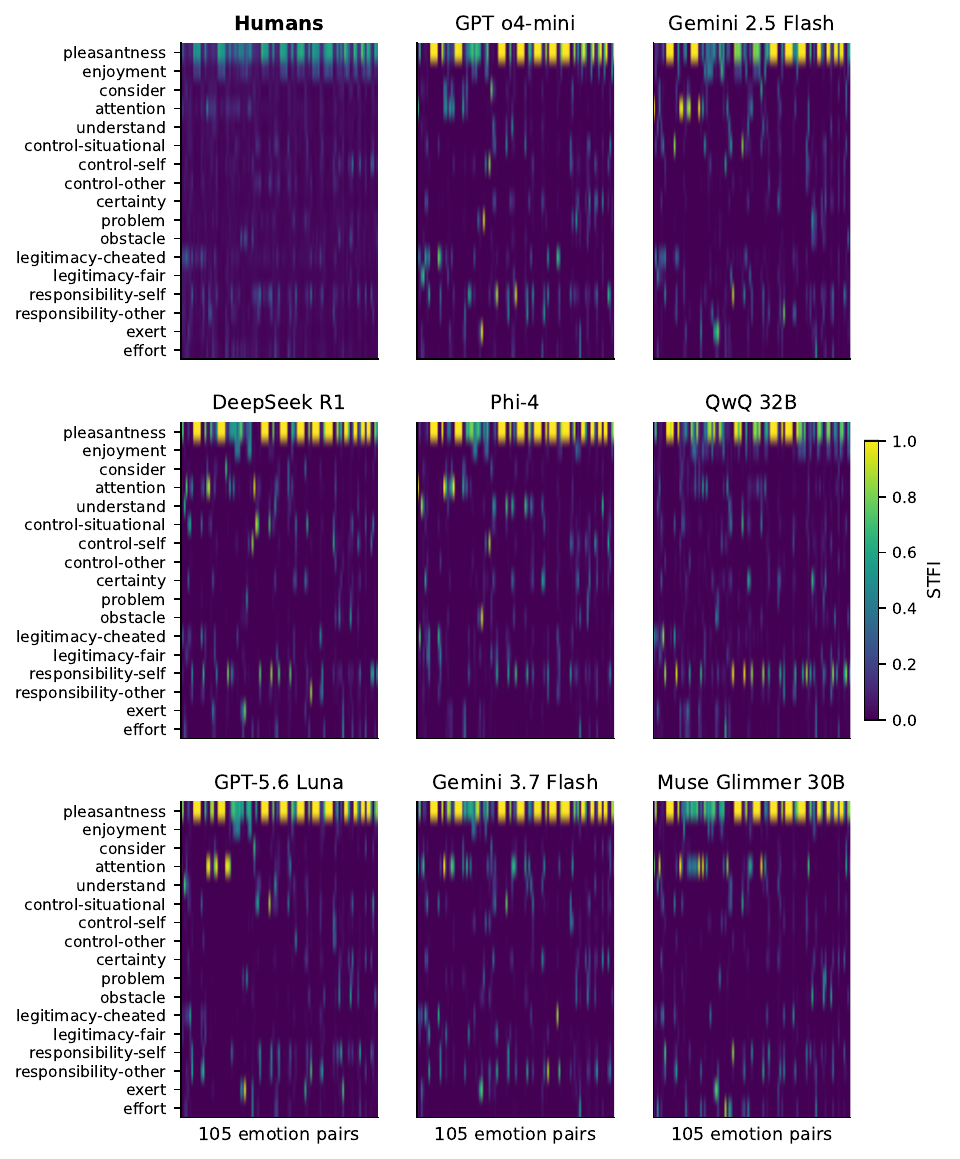}
  \caption{Appraisal-dimension importance across all emotion pairs. For each sub-plot, representing a single agent, each column denotes the $105$ possible emotion pairs, and is left unlabelled for brevity. Discriminators fitted on LLMs' behavioral representations places most of its importance on fewer dimensions for any given pair, compared to humans. Table~\ref{tab:synthtree_summary} reports the same contrast numerically, as the concentration of the importance profile.}
  \label{fig:stfi-heatmap}
\end{figure}

\paragraph{Discriminators in LLMs' behavioral representations rely on fewer appraisal dimensions than for humans.} We compute the importance assigned to a given appraisal dimension by the discriminator fitted on each agent's appraisal space. For LLMs, the importance is concentrated on fewer appraisal dimensions than humans: the Gini coefficient of the importance vector is $0.55$ for humans and
between $0.81$ and $0.86$ for the models (Table~\ref{tab:synthtree_summary}), and the same contrast is visualized in detail in Figure~\ref{fig:stfi-heatmap}. The difference between humans' and LLMs' behavioral representations is, however, in how concentrated the attributed importance is, and they do not necessarily differ along \textit{which specific dimension} is assigned the most importance. For both humans and models, the single most important dimension is usually pleasantness: it ranks first in 65 of the 105 human discriminators and in 463 of the 840 model discriminators. However, in 154 of the 840 model discriminators, all of the importance rests
on a single dimension alone, while this cannot be observed for any of the human discriminators. This difference is a property of the ratings rather than of the fitting procedure. For each pair, the rank correlation between the fitted importance profile and the per-dimension discriminability of the ratings themselves, measured as the absolute Cliff's delta between the two emotions, is $0.58$ for
humans and between $0.58$ and $0.68$ for the models
(Table~\ref{tab:synthtree_summary}), so the discriminator tracks the ratings to the same degree for every agent.

\begin{figure}[p]
  \centering
  \includegraphics[scale=0.787]{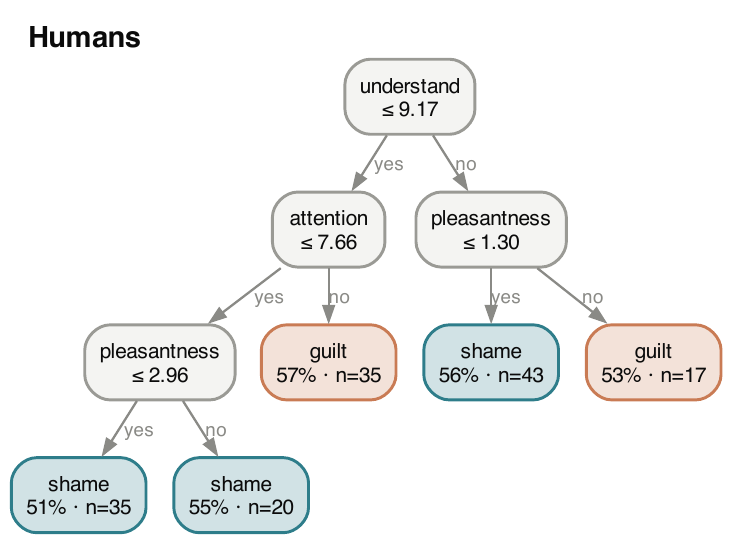}\\[0.25cm]
  \raisebox{-\height}{\includegraphics[scale=0.787]{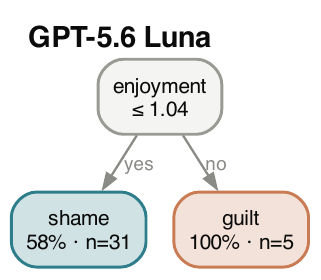}}\hspace{0.40cm}%
  \raisebox{-\height}{\includegraphics[scale=0.787]{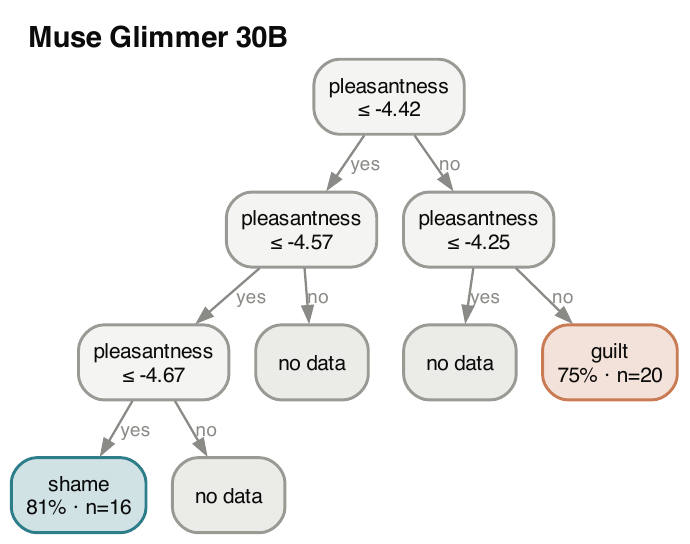}}
  \caption{Fitted discriminators for \emph{guilt} against \emph{shame}, for: humans, GPT-5.6 Luna and Muse Glimmer 30B, each labelled
  inside the respective sub-plot. A split node shows the appraisal dimension it tests and the
  threshold it tests at. A leaf reports the majority true class of the training rows routed to it, that
  class's share of the leaf, and the corresponding number of rows. A leaf that no training row
  reaches is shown in grey. The trees for visualization are fit on all rows across the entire data comparing these two emotions, and not cross-validated. The percentages here thus describe the data fit rather than accuracy on
  held-out data.}
  \label{fig:synthtree-trees}
\end{figure}

\paragraph{Visualization of discriminators.} Figure~\ref{fig:synthtree-trees} shows the form of the fitted discriminators for guilt against shame, for the human ratings and for two models, GPT-5.6 Luna and Muse Glimmer 30B. The pair shown is also where the models show relatively lower cross-validated balanced accuracy ($0.55$ for humans and between $0.56$ and $0.85$ for the models). 
 
 
\paragraph{Correspondence with the agents' stated discriminators.} For each agent we compare the STFI profile of the fitted discriminator with the
agent's explicit statements on which dimensions separate the pair, as a rank correlation over the 17 dimensions (similar to the results shown in main body and Appendix~\ref{sec:app_implicit_explicit}). Computing the correlations for the same subsampled 90 emotion pairs as used in the main implicit-explicit results and Appendix~\ref{sec:app_implicit_explicit}, we find that the correlation is between $0.28$ and $0.43$ for the eight models and $0.04$ for humans (Table~\ref{tab:synthtree_summary}). This independently replicates the results we already obtain in the main body and in Appendix~\ref{sec:app_implicit_explicit}). However, we interpret these results to only show that the determinacy of emotion concepts in LLMs is higher in comparison to humans. We do not attribute a direct cause of the low implicit-explicit correspondence in humans, as it could also stem from a lack of introspective access in making emotional judgements~\citep{schwitzgebel2008unreliability}. 


\begin{table}
\centering
\small
\setlength{\tabcolsep}{6pt}
\resizebox{\textwidth}{!}{
\begin{tabular}{lcccc}
\toprule
Agent & \shortstack{SynthTree acc.} & \shortstack{STFI concentration} & \shortstack{$\rho$(STFI, explicit)} & \shortstack{$\rho$(STFI, implicit)} \\
\midrule
Humans & 0.76 & 0.55 & 0.04 & 0.58 \\
\midrule
GPT o4-mini & 0.96 & 0.84 & 0.40 & 0.63 \\
Gemini 2.5 Flash & 0.96 & 0.83 & 0.43 & 0.67 \\
DeepSeek R1 & 0.97 & 0.86 & 0.42 & 0.58 \\
Phi-4 & 0.96 & 0.84 & 0.28 & 0.59 \\
QwQ 32B & 0.94 & 0.81 & 0.29 & 0.68 \\
GPT-5.6 Luna & 0.97 & 0.86 & 0.31 & 0.61 \\
Gemini 3.7 Flash & 0.97 & 0.86 & 0.43 & 0.64 \\
Muse Glimmer 30B & 0.97 & 0.86 & 0.38 & 0.61 \\
\bottomrule
\end{tabular}}
\caption{Interpretable discriminators of emotion pairs, shown per agent. SynthTree acc.: mean balanced accuracy on held-out folds. STFI concentration: the Gini coefficient of the importance the fitted discriminator assigns to the 17 appraisal dimensions, averaged over pairs, where a higher value means the discriminator rests on fewer dimensions.  $\rho$(STFI, explicit): the rank correlation between the importance profile and
the agent's own statement of which dimensions separate the pair. This is computed across the 90 psychologically-informed, subsampled emotion pairs. $\rho$(STFI, implicit): the rank correlation between that importance profile and the per-dimension discriminability of the ratings themselves. }
\label{tab:synthtree_summary}
\end{table}

\clearpage
{\scriptsize\setlength{\tabcolsep}{2pt}
{\raggedright\noindent\textit{Abbreviated dimensions:} ctrl-situ~=~control-situational; ctrl-self~=~control-self; ctrl-other~=~control-other; legit-cheat~=~legitimacy-cheated; legit-fair~=~legitimacy-fair; resp-self~=~responsibility-self; resp-other~=~responsibility-other.\par}\medskip
\begin{longtable}{llccclll}
\toprule
Pair & Agent & Acc. & \shortstack{Bal.\\acc.} & \shortstack{Macro\\F1} & Top 1 & Top 2 & Top 3 \\
\midrule
\endfirsthead
\toprule
Pair & Agent & Acc. & \shortstack{Bal.\\acc.} & \shortstack{Macro\\F1} & Top 1 & Top 2 & Top 3 \\
\midrule
\endhead
\midrule
\multicolumn{8}{r}{Continued on next page} \\
\midrule
\endfoot
\bottomrule
\endlastfoot
anger vs boredom & Humans & 0.69 & 0.69 & 0.69 & legit-cheat & attention & legit-fair \\
anger vs boredom & GPT o4-mini & 0.97 & 0.97 & 0.97 & attention & legit-cheat & pleasantness \\
anger vs boredom & Gemini 2.5 Flash & 1.00 & 1.00 & 1.00 & attention & consider & legit-cheat \\
anger vs boredom & DeepSeek R1 & 0.98 & 0.97 & 0.98 & attention & pleasantness & ctrl-situ \\
anger vs boredom & Phi-4 & 1.00 & 1.00 & 1.00 & attention & legit-fair & exert \\
anger vs boredom & QwQ 32B & 0.94 & 0.94 & 0.94 & legit-cheat & pleasantness & legit-fair \\
anger vs boredom & GPT-5.6 Luna & 1.00 & 1.00 & 1.00 & attention & pleasantness & consider \\
anger vs boredom & Gemini 3.7 Flash & 0.97 & 0.98 & 0.97 & attention & legit-cheat & pleasantness \\
\end{longtable}
}

\captionof{table}{Pairwise held-out accuracy, denoting separability of emotion pairs, and the three most important appraisal dimensions. Acc.\ and macro $F_1$ are computed on held-out folds under repeated stratified cross-validation. Top~1 to Top~3 provide the three dimensions with the largest
feature importance for that agent on that pair, in order. An em-dash marks a rank whose importance is exactly zero, which happens when the discriminator
places all of its importance on a single dimension, so that no second or third dimension exists to name.}
\label{tab:synthtree_pairs}


\section{Convergence across Agents}
\label{sec:app_convergence}

\paragraph{Details of the Method.} As stated in the main body, our goal in this section is to understand whether, beyond discretization and homogeneous representations produced behaviorally by an \textit{individual} LLM, do distinct models also show higher concurrence of the representations compared to distinct groups of humans? To do so, we measure the multivariate Wasserstein distance between the complete representations (of dimension $n \times 17$, where $n$ denotes the number of scenarios for the given emotion) for a given emotion category from two different LLMs (out of 8 total models). For humans, as no single rater rates all scenarios across an emotion, we approximate the population by dividing the participants into 8 random groups, where ratings per scenario are averaged within such a group. This enables the same kind of ($8\choose{2}$ $=28$) pairwise comparisons for both humans and LLMs. The random sampling process is repeated 300 times.

\paragraph{Notes on Qualitative Trends in the Inter-Agent Convergence Results.} Inter-model and human agreement show similar qualitative trends. Different human groups agree the most on positive emotions such as happiness, pride, and interest, while agreeing the least on emotions like sadness, fear, and disgust. Inter-LLM agreement is also qualitatively similar, with high agreement on emotions including pride and happiness, and low agreement on disgust and fear. Models further show high inter-agent agreement on the relatively nuanced emotion of challenge~\citep{uphill2019challenge}. 

\paragraph{Distance-free measure of convergence.} In our main results, we show that LLMs from different sources have significantly lower distributional distances than random groups of human participants. We further explore a distance-free measure of inter-agent similarity. For each agent and each emotion category, we average that source's ratings along a single appraisal dimension, across all scenarios of that emotion. This yields one value
per appraisal dimension, for a given emotion category. We then replace those values by their ranks, retaining only the order of appraisal dimensions per emotion category. Kendall's coefficient of concordance $W$ is then used to calculate the agreement of the same ordering across different agents. For LLMs, a single model is considered a different agent, while for humans, we use the same process of randomly dividing participants into 8 groups, repeating the process $300$ times. This provides a measure of agreement that does not depend on a specific distance measure. As in all main analyses, ranks are calculated after standardizing ratings within each appraisal dimension. 

Averaged over the fifteen emotions, concordance among the eight models is
$W=0.895$, against $0.617$ among the human groups. Models are more concordant
for every emotion without exception, and in every case the model value lies
above the upper end of the human $95\%$ range, so none of the $300$ random
divisions of the participants produced agreement as high as the models'
($p<0.0033$ for each emotion). Model concordance is high and narrowly spread, from
$0.822$ for boredom to $0.957$ for interest. Human concordance is both lower
and far more variable, from $0.378$ for disgust to $0.837$ for happiness. The
largest gaps are for disgust ($0.846$ against $0.378$), sadness ($0.877$ against
$0.501$) and boredom ($0.822$ against $0.436$), whereas the smallest is happiness
($0.940$ against $0.837$), the one emotion on which people also broadly agree. These results arrive at the same quantitative conclusion, that LLMs are convergent across different families, and show the same qualitative trends.

\paragraph{Introducing sampling noise in models.} 
Similar to all of our main analyses, we consider here the effect of injecting measurement error/sampling noise into the model ratings. For one emotion, the inter-agent distance is the mean $W_2$ distance between the appraisal profiles of two agents, averaged over the 28 pairs of the eight models. The human reference is built the same way: all participants are divided at random into eight groups, the ratings are averaged
within each group for each scenario, and the 28 distances between the groups are computed. We repeat this random partitioning and division 300 times and report the median and the $2.5$th and $97.5$th percentiles. As for the previous analyses, we add Gaussian noise to the model ratings so that they carry the same measurement error as the human ratings, with the human data left unchanged. We use five independent noise draws and report the mean.

Table~\ref{tab:noise_conv} reports the result. Averaged over the fifteen emotions, the distance between two distinct models increases from $2.37$ to $3.76$ under matched error, but remains lower than a distance of $4.14$ between two human groups. The models remain closer to one another than the human groups are on all fifteen emotions, both before and after matching (paired Wilcoxon signed-rank test over the fifteen emotions, $p < 0.001$ in both cases). The core finding thus holds even under the stricter comparison of noise-matched model ratings. 

\begin{table}[t]
\centering
\small
\setlength{\tabcolsep}{5pt}
\begin{tabular}{@{}lccc@{}}
\toprule
 & \multicolumn{2}{c}{Between two models} & Between two \\
\cmidrule(lr){2-3}
Emotion & Default & Noise-matched & human groups \\
\midrule
anger       & 1.95 & 3.62 & 3.97 [3.76, 4.19] \\
boredom     & 2.68 & 3.86 & 4.08 [3.82, 4.31] \\
challenge   & 1.94 & 3.40 & 3.95 [3.73, 4.15] \\
contempt    & 2.80 & 3.93 & 4.30 [4.06, 4.50] \\
disgust     & 3.19 & 4.39 & 4.59 [4.31, 4.85]  \\
fear        & 2.85 & 4.25 & 4.41 [4.16, 4.61]  \\
frustration & 2.09 & 3.60 & 4.05 [3.84, 4.31] \\
guilt       & 2.23 & 3.55 & 3.98 [3.76, 4.20] \\
happiness   & 1.69 & 3.39 & 3.82 [3.55, 4.07]  \\
hope        & 2.24 & 3.70 & 4.02 [3.75, 4.25] \\
interest    & 2.07 & 3.40 & 3.89 [3.67, 4.12] \\
pride       & 1.94 & 3.44 & 3.84 [3.61, 4.09]  \\
sadness     & 2.51 & 4.03 & 4.79 [4.53, 4.99]  \\
shame       & 2.88 & 4.01 & 4.12 [3.90, 4.33]  \\
surprise    & 2.44 & 3.80 & 4.31 [4.07, 4.51]  \\
\midrule
Mean        & 2.37 & 3.76 & 4.14               \\
\bottomrule
\end{tabular}
\caption{Distance between the emotion representations of two agents, with the
model ratings given the same measurement error as the human ratings. Between two
models: the mean $W_2$ distance over the 28 model pairs. Between two human
groups: the median over 300 random divisions of the participants into eight
groups, with the $2.5$th and $97.5$th percentiles in brackets.}
\label{tab:noise_conv}
\end{table}

\paragraph{Cloud size and scenario coverage.} We further note that the original convergence comparison has an additional point of asymmetry: a model rates every scenario, so a model's cloud for an emotion contains all of that emotion's scenarios, on average $17.4$ of them. A human group of five participants covers only the scenarios its own members rated, on average $6.6$. The empirical Wasserstein distance between two small clouds is biased upward, and two human groups could also differ because they cover different scenarios rather than because they judge them differently. Both effects work in the direction of the reported result, so we individually test each of them. In the first control, each model's cloud is reduced at random to the number of scenarios in the matched human group's cloud, so the two sides are compared at the same cloud size. In the second, each pair of human groups is restricted to the scenarios both groups cover, which removes the difference in coverage. Both controls use the same $300$ random divisions of the participants as the main analysis, and everything else remains unchanged from the original comparison.

Table~\ref{tab:conv_cloudsize} reports the result. Reducing the model clouds to the size of the human clouds raises the distance between two models from $2.38$ to $2.77$, and the ratio between the human and the model distance falls from $1.74$ to $1.50$, showing that a very small part of the reported gap stems from differing cloud sizes. Restricting the human groups to their shared scenarios moves the human distance in the opposite direction, from $4.15$ to $4.43$: the shared subset contains only about four scenarios, and the further size reduction outweighs what is gained by removing the difference in coverage. Applying both controls together yields $2.94$ against $4.43$, a ratio of $1.51$. Under \textit{all} three controls, the models remain significantly closer to one another than the human groups are, on every one of
the fifteen emotions, with the same paired Wilcoxon signed-rank result as the main analysis ($p<0.001$). 
Finally, giving each model exactly the scenarios its matched human group rated, rather than a random subset of the same size, yields the same result to two decimal
places ($2.76$ against $2.77$). The differing cloud size thus accounts for a small part of the size of the human-LLM difference, and none of its direction.

\begin{table}[t]
\centering
\small
\setlength{\tabcolsep}{5pt}
\begin{tabular}{@{}lcccc@{}}
\toprule
 & \multicolumn{2}{c}{Between two models}
 & \multicolumn{2}{c}{Between two human groups} \\
\cmidrule(lr){2-3}\cmidrule(lr){4-5}
Emotion & Full clouds & Size-matched & All scenarios & Shared scenarios \\
\midrule
anger       & 1.96 & 2.17 & 3.95 & 4.27 \\
boredom     & 2.70 & 3.04 & 4.11 & 4.50 \\
challenge   & 1.94 & 2.27 & 4.04 & 4.26 \\
contempt    & 2.82 & 3.03 & 4.33 & 4.68 \\
disgust     & 3.17 & 3.71 & 4.52 & 4.79 \\
fear        & 2.87 & 3.39 & 4.43 & 4.77 \\
frustration & 2.10 & 2.42 & 4.07 & 4.28 \\
guilt       & 2.24 & 2.49 & 3.98 & 4.40 \\
happiness   & 1.72 & 2.18 & 3.85 & 4.17 \\
hope        & 2.25 & 2.70 & 4.01 & 4.35 \\
interest    & 2.09 & 2.52 & 3.89 & 4.24 \\
pride       & 1.95 & 2.39 & 3.88 & 4.07 \\
sadness     & 2.51 & 3.02 & 4.74 & 4.97 \\
shame       & 2.93 & 3.27 & 4.14 & 4.32 \\
surprise    & 2.45 & 2.89 & 4.32 & 4.43 \\
\midrule
Mean        & 2.38 & 2.77 & 4.15 & 4.43 \\
\bottomrule
\end{tabular}
\caption{Distance between the emotion representations of two agents, under two
controls for the number of scenarios in a cloud. Full clouds: the mean $W_2$
distance over the 28 model pairs, as in Table~\ref{tab:noise_conv}.
Size-matched: each model's cloud reduced at random to the size of the matched
human group's cloud. All scenarios: the median over 300 random divisions of the
participants into eight groups, as in Table~\ref{tab:noise_conv}. Shared
scenarios: the same divisions, with each pair of groups restricted to the
scenarios both groups cover. Model clouds hold $17.4$ scenarios on average,
human group clouds $6.6$, and the shared subset of two human groups $4.0$.}
\label{tab:conv_cloudsize}
\end{table}

\section{Responses without an Emotion Label}
\label{sec:app_without_emotion}

\paragraph{Measuring emotion fidelity.} As a natural follow-up, we experiment with what happens with the specific emotion label is removed from the prompt. This serves to identify whether the observed discretization is a property of only the label being present in the prompt, and in the main results, we show that it slightly changes the between- and within-emotion distances, but model representations still remain significantly different from humans. 

This however introduces the possibility that a model may appraise a situation as a different emotion from the one the scenario was designed to elicit. To understand whether this happens, we use the free-text response generated by the model, and use an LLM-as-a-judge~\citep{li2025generation} to map the responses to one of the 15 available emotion categories. To do this, we employ GPT-OSS 120B~\citep{agarwal2025gpt}, a strong, open-source model, which we query through the OpenRouter platform (\texttt{https://openrouter.ai/}). Further, as we sample responses for each appraisal dimension independently from a given model, the emotion response to each scenario is output by a single model $n=17$ times. All 17 such descriptions are shown together in a single call to the LLM judge, with the instruction that they were written by one person about one situation on separate occasions. The judge is also instructed to choose
choose one category from the list of 15 emotion categories, may not refuse or invent a category,
and also returns a confidence from 1 to 5 and a reason of at most fifteen words, in a fixed JSON form. The judge is not shown the scenario, so it cannot recover the intended emotion from the situation itself, it sees only the evaluated model's report. Decoding is deterministic
by setting temperature at 0. Replies that do not resolve to one of the fifteen categories
are counted as incorrect, and are 0.1\% of responses judged across all models. A
scenario counts as correctly interpreted when the judge's category is the emotion the scenario was written to elicit. We present a preview of what the exact judge prompt looks like in practice, as follows:
\begin{promptbox}[LLM-as-a-judge Evaluation of Interpreted Emotion]

\textcolor{cyan!80}{----- system -----} \\
You are an expert annotator of emotional language. You read what a person says about how they felt and identify which single emotion category best describes it. You answer only in the requested JSON format. \\

\textcolor{cyan!80}{----- user prompt -----} \\
Below are statements written by one person about how they felt in one particular situation. The statements were written on separate occasions about the same situation, so they may differ in wording. \\

Statements: 
\begin{enumerate}
    \item I felt a surge of pure joy as I savored each bite.
    \item I felt an overwhelming sense of joy when I ate my favorite food.
    \item I felt an overwhelming sense of joy as I savored every bite of my favorite food.
    \item I feel joyful and content after eating my favorite food.
    \item I felt pure joy and contentment.
    \item I felt an overwhelming sense of happiness as I savored each bite of my favorite meal.
    \item I felt pure bliss savoring every bite of my favorite food.
    \item I felt an overwhelming sense of joy and contentment.
    \item I felt an overwhelming sense of joy and satisfaction.
    \item I felt a wave of pure joy and contentment.
    \item I felt pure delight and satisfaction as I enjoyed my favorite food.
    \item I feel joyful and content after enjoying my favorite food.
    \item I felt a burst of pure happiness and satisfaction as I enjoyed my favorite dish.
    \item I felt an overwhelming sense of joy and contentment after savoring my favorite food.
    \item I felt a surge of pure joy and contentment.
\end{enumerate}

Which one of the following fifteen emotion categories best describes the emotion these statements express? \\

\begin{itemize}
    \item anger
    \item frustration
    \item contempt
    \item fear
    \item disgust
    \item boredom
    \item sadness
    \item guilt
    \item shame
    \item challenge
    \item surprise
    \item hope
    \item interest
    \item pride
    \item happiness
\end{itemize}
  
Choose exactly one category. If more than one seems to fit, choose the one that fits best; do not refuse and do not name a category outside the list. Then give your confidence in that choice on a scale from 1 (a guess) to 5 (certain), and a brief reason of at most fifteen words. \\

Answer in JSON, in exactly this form:
{"emotion": "<one category from the list>", "confidence": <1-5>, "reason": "<brief reason>"}    
\end{promptbox}
 
\paragraph{Results of emotion fidelity.} Table~\ref{tab:nolabel_correct}
reports, for each model, how many of the total scenarios were interpreted as the intended emotion, based on the evaluation by our LLM judge. The share ranges from $0.46$ for Gemini 2.5 Flash to $0.57$ for Gemini 3.7 Flash and Muse Glimmer 30B, against $0.067$ for
choosing at random among the fifteen categories, and the eight models average $0.53$. The judge is confident in these readings, at $4.7$ to $4.9$ out of 5 on average. Which emotions are recovered is nearly the same for every model. Six
categories are read back correctly at least three quarters of the time, namely
pride, sadness, guilt, fear, disgust and frustration; four are read back rarely
or never, namely contempt (no model, on any scenario), challenge ($0.00$ to
$0.12$), anger ($0.00$ to $0.36$) and boredom ($0.00$ to $0.47$). The misplaced
scenarios do not scatter: frustration absorbs the largest number in every model
(24 to 47 scenarios), followed by happiness, sadness and anger, while contempt
attracts none. Without the label, then, a model often writes about a situation
as a neighbouring emotion, and the neighbour it chooses is shared across models.
 
\paragraph{Separation restricted to the correctly interpreted scenarios.} A
model that construes a scenario as a different emotion is, in effect, being
scored on the wrong category, so the separation measured on all scenarios may
understate how sharply the model separates the emotions it actually
represents. We therefore recompute the between-emotion distance, the
within-emotion spread and their ratio on the subset of scenarios the judge read
back as the intended emotion, which is a different subset for each model and
contains 121 to 150 scenarios. Emotions left with fewer than three scenarios in
that subset are dropped, which removes contempt and challenge for every model
and, for Gemini 2.5 Flash, also boredom, hope and interest. The human ratings
are recomputed on the same subset as the model they are compared with, so the
two are always measured on the same scenarios. Because the subset is both
smaller and easier, we add a control: for each model we draw twenty random
subsets with the same number of scenarios per emotion as its correct subset and
average the result, which shows what the reduction in size alone does.
Table~\ref{tab:nolabel_correct} reports all three. Restricting to correctly
interpreted scenarios does not move the models towards humans. On its own
correct subset each model's ratio is $1.51$ to $1.85$, against $1.24$ to $1.33$
for the human ratings on the same scenarios, so the difference between models
and people is of the same size as on the full set. The apparent fall for some
models from the full set to the correct subset, largest for Gemini 2.5 Flash at
$2.01$ to $1.52$, is a consequence of the smaller sample rather than of the
selection: the random subsets of the same size give $1.42$ to $1.72$, which
brackets the correct-subset values for every model, and the human ratio falls in
the same way. What changes with the subset is the within-emotion spread, which
rises for models and for humans alike as fewer scenarios are averaged; the
between-emotion distance is almost unchanged. The conclusion of the
without-label analysis is therefore unaffected by how often a model interprets
the scenario as intended.
 
 
\begin{table}[t]
\centering
\small
\setlength{\tabcolsep}{5pt}
\resizebox{\textwidth}{!}{
\begin{tabular}{@{}lcccccccc@{}}
\toprule
 & & \multicolumn{4}{c}{Correctly interpreted scenarios}
 & \multicolumn{1}{c}{Same size} & \multicolumn{2}{c}{All scenarios} \\
\cmidrule(lr){3-6}\cmidrule(lr){7-7}\cmidrule(lr){8-9}
Model & $\%$ & Between & Within & Ratio & Humans & Ratio & Ratio & Humans \\
\midrule
Muse Glimmer 30B & 57\% & 5.24 & 2.84 & 1.85 & 1.33 & 1.70 & 1.84 & 1.31 \\
Gemini 3.7 Flash & 57\% & 5.29 & 2.89 & 1.83 & 1.32 & 1.72 & 1.87 & 1.31 \\
GPT-5.6 Luna     & 55\% & 5.29 & 2.93 & 1.81 & 1.27 & 1.71 & 1.87 & 1.31 \\
DeepSeek R1      & 53\% & 5.13 & 3.22 & 1.60 & 1.25 & 1.48 & 1.57 & 1.31 \\
Phi-4            & 49\% & 4.93 & 3.13 & 1.57 & 1.24 & 1.52 & 1.66 & 1.31 \\
GPT o4-mini      & 52\% & 5.14 & 3.37 & 1.53 & 1.24 & 1.55 & 1.65 & 1.31 \\
Gemini 2.5 Flash & 46\% & 4.76 & 3.13 & 1.52 & 1.24 & 1.57 & 2.01 & 1.30 \\
QwQ 32B          & 55\% & 5.16 & 3.42 & 1.51 & 1.27 & 1.42 & 1.47 & 1.31 \\
\bottomrule
\end{tabular}}
\caption{Separation of emotion concepts in the without-label condition, on the
scenarios each model interpreted as the intended emotion. Between: the mean
$W_2$ distance between two emotion categories; Within: the mean spread inside
one category; Ratio: their quotient. Humans: the human ratio computed on that
model's own subset. Same size: the mean ratio over twenty random subsets with
the same number of scenarios per emotion, which isolates the effect of the
smaller sample. All scenarios: the same quantities on every scenario the model
rated without the label.}
\label{tab:nolabel_correct}
\end{table}

\section{Responses from a Third-person Perspective}
\label{sec:app_third_person}

\paragraph{Details of the Setup.} We consider the effect of the prompt framing, testing whether posing the questions with a third-person framing leads to a change in the representation structure. Within the same set of prompts, we use the persona framing of \textit{"Consider a person who..."}, replacing \textit{"Imagine you are a person who..."}. 

\paragraph{Full qualitative results.} In this section, we provide additional detailed results on the qualitative changes in appraisal distributions, when responses are elicited using the third-person perspective. To further characterize whether the appraisal responses change qualitatively with this framing, we compare each model's first- and third-person response distributions at three levels of granularity, with a distributional two-sample test at each level. First, for each appraisal dimension, we pool responses across all emotions and test whether the dimension's distribution changes between perspectives, using the maximum mean discrepancy (MMD) with a permutation test. Second, for each emotion, we pool across all appraisal dimensions and test whether the emotion's joint, multivariate appraisal profile changes, standardizing each dimension before computing a multivariate MMD so that dimensions on larger numeric scales do not dominate. Because these two levels ask only whether a distribution shifts---a multivariate change has no single direction---we report them without a directional sign. Third, for each (emotion $\times$ appraisal) pair, where a change is one-dimensional and its direction is interpretable, we test for a difference with the Mann–Whitney U test and quantify its size and sign with Cliff's $\delta$, where a positive value indicates higher ratings under the first-person framing. 

\paragraph{Perspective-sensitivity is model-dependent.}
As we show in our main results, the geometry of how emotions are represented remains largely preserved under a third-person framing. We find, however, that the ratings themselves are not perspective-invariant, and the extent of the
shift differs across models. At the level of individual appraisal dimensions across all emotions, GPT o4-mini shows \textit{no dimension} that changes significantly (after FDR correction), and Gemini 3.7 Flash and QwQ 32B change on only one each (\emph{attention} and \emph{effort}, respectively). Appraisals from Gemini 2.5 Flash change along three dimensions (\emph{attention}, \emph{understanding} and \emph{certainty}) and GPT-5.6 Luna on four, while DeepSeek R1 ratings changes along five dimensions, and Phi-4 and Muse Glimmer 30B six each.

The extent of change holds at the level of an emotion's full multivariate appraisal representation: no emotion changes significantly for GPT o4-mini, Gemini 2.5 Flash, QwQ 32B or GPT-5.6 Luna, exactly one emotion changes for DeepSeek R1, Phi-4 and Gemini 3.7 Flash (\emph{contempt} in all three), whereas for Muse
Glimmer 30B eight of the fifteen emotions change (\emph{anger}, \emph{challenge}, \emph{contempt}, \emph{frustration}, \emph{guilt}, \emph{interest}, \emph{sadness} and \emph{shame}).
Finally, across all 255 (emotion $\times$ appraisal) pairs, the number of significant changes ranges from 0 (GPT o4-mini, Gemini 2.5 Flash, QwQ 32B and GPT-5.6 Luna) and 1 (Gemini 3.7 Flash) through 10 (DeepSeek R1 and Phi-4) to 38 (Muse Glimmer 30B). Perspective-sensitivity is thus concentrated in a few models. 

\paragraph{Negative situations are appraised as less demanding and less aversive from the outside.} We find that the significant changes at the level of a given appraisal-emotion pair are concentrated on the effort dimensions and on negatively valenced emotions. All significant changes along the \emph{effort}/\emph{exertion} dimensions ($21$ total such pairs, across all emotions and models)
are higher under first-person framing. This indicates that models judge the same situation to demand less effort when the task is framed as third-person appraisal. The emotion categories for which this is observed are all negatively valenced: \emph{contempt} ($\delta=0.77$ effort, $0.69$ exertion in DeepSeek R1; $0.91$ effort in Gemini 3.7 Flash; $0.68$ effort and $0.72$ exertion in Muse Glimmer 30B; $0.68$ effort in Phi-4), \emph{guilt} ($0.87$ effort and $0.57$ exertion in Muse Glimmer 30B; $0.62$ effort in Phi-4; $0.59$ effort and $0.57$ exertion in DeepSeek R1), \emph{shame} ($0.76$ effort and $0.77$ exertion in Muse Glimmer 30B; $0.60$ effort in Phi-4; $0.58$ exertion in DeepSeek R1), \emph{anger} ($0.74$ effort and $0.49$ exertion in Muse Glimmer 30B; $0.62$ effort in DeepSeek R1), \emph{disgust} ($0.56$ effort and $0.58$ exertion in Muse Glimmer 30B), \emph{fear} ($0.51$ effort in Muse Glimmer 30B) and \emph{sadness} ($0.63$ exertion in Muse Glimmer 30B).
A parallel change is observed along valence-related dimensions. Negative scenarios are rated as \emph{more} pleasant when appraised in a third-person setting---\emph{contempt} in DeepSeek R1 ($\delta=-0.64$), and \emph{contempt} ($-0.53$) and \emph{anger} ($-0.32$) in Muse Glimmer 30B, which also rates \emph{contempt} as more enjoyable from the outside ($-0.47$).
In Muse Glimmer 30B, further, perception of unfairness (being cheated) falls markedly under third-person framing across \emph{boredom} ($\delta=0.56$), \emph{contempt} ($0.51$) and
\emph{guilt} ($0.44$). These shifts describe a consistent softening of negative appraisal in a third-person framing: the observer's version of a distressing situation is appraised as less effortful, less unpleasant, and less unfair than the first-person version.

\paragraph{Epistemic and agency dimensions shift, but inconsistently across models.}
Beyond the above patterns, perspective also reshapes appraisals of comprehension and
control, albeit in a model-dependent manner. Phi-4 reports \emph{lower} understanding of a situation experienced by someone else
(\emph{happiness} $\delta=0.65$, \emph{interest} $0.61$, \emph{contempt} $0.57$), whereas Muse Glimmer 30B reports \emph{higher} understanding in the same condition, and does so across seven emotions (\emph{guilt} $\delta=-0.91$, \emph{shame} $-0.82$, \emph{frustration} $-0.76$, \emph{fear} $-0.63$, \emph{sadness} $-0.62$, \emph{disgust} $-0.59$, \emph{anger} $-0.57$). Agency attributions move in a single direction but are confined to two models and fewer emotions, pointing to the lack of a more general pattern: DeepSeek R1 and Muse Glimmer 30B both grant the observed agent less situational control (\emph{challenge} $\delta=0.53$ and \emph{shame} $0.58$ in DeepSeek R1; \emph{challenge} $0.63$ in Muse Glimmer 30B, which also lowers other-control for \emph{challenge}, $0.48$), DeepSeek R1 lowers other-responsibility for the observed agent in \emph{guilt} ($0.67$), and certainty falls for \emph{surprise} in Muse Glimmer 30B ($0.66$) and for \emph{pride} in Phi-4 ($0.63$). That the effort and aversiveness shifts are directionally consistent across every model in which they appear, while the comprehension shifts reverse
between models, suggests the former reflect a more systematic property of how these models represent the experience of others, whereas the latter are model-specific.

In summary, the observed rating-level changes are real but bounded. They alter qualitatively \textit{where} an emotion is placed in the appraisal space without altering the overall discretized structure of the space. The changes are also largely confined to a handful of dimensions in three of the eight models and
absent altogether in four of them. Adopting an observer's perspective therefore modulates the content of a model's appraisals without altering the determinacy of its emotion representations, and does not recover the scenario-to-scenario variability that characterizes the human data.

\section{Responses with Cultural Context}
\label{sec:app_culture}

\paragraph{Details of the Setup. }Given widespread evidence of nationality-based personas inducing variability in LLMs in terms of cultural disposition~\citep{alkhamissi2024investigating, kamruzzaman2025anger}, we test whether inducing cultural personas shifts the homogeneity of emotion representations in models. We use nationality as a proxy for culture, as it provides a succinct, abstracted, and effective way of inducing culturally variant responses~\citep{kwok2024evaluating, alkhamissi2024investigating}. We include nationality-based personas corresponding to USA, Mexico, Nigeria, Denmark, and Japan. Exact prompts are in Appendix~\ref{sec:app_prompt_templates}. We choose these countries to represent major geopolitical regions of the world and to ensure diversity along several dimensions of Hofstede's theory of cultures \citep{hofstede1984culture}. This also aligns with the intuition that maximally different cultures may exhibit large differences in cognitive appraisals of emotion. For example, Japan and the USA are on opposite ends of the individualism--collectivism spectrum, whereas Denmark and Latin American cultures (e.g., Mexico) are on opposite ends of the Power Distance Index spectrum. Thus, five cultural personas lead to the creation of 22,950 (= vanilla prompts (4,590) $\times$ 5) additional prompts as part of CoRE. 

\paragraph{Full Qualitative Results.} We find that the structural invariance, when nationality-based personas are used, also extends to the qualitative nature of the appraisals. Using the Maximum Mean Discrepancy (MMD) metric, we find, surprisingly, that appraisal representations from all LLMs remain statistically indistinguishable with and without any cultural persona. This denotes that culture-specific prompting, although found to be effective in the past for normative alignment~\citep{alkhamissi2024investigating}, does not influence nuanced cultural \textit{disposition}. 

\section{Responses with Personality-based Context}
\label{sec:app_personality}

\paragraph{Details of the Setup. }Along with culture, we also examine whether different personality traits impact the appraisal-based representations of emotions in LLMs. Studies in humans have repeatedly recorded how different personality traits impact appraisals of emotion-related stimuli, often propagating downstream into actions such as coping or recognizing threat~\citep{gunthert1999role,sacchi2026understanding}. We use prompts from an existing study, based on the Big Five personality traits, to provide personality-based personas in-context~\citep{liu2025synthetic}. For each attribute from the Big Five traits---Agreeableness, Conscientiousness, Extraversion, Neuroticism, and Openness---two personas are created, embodying either a strong presence or an absence of the attribute. For example, for Agreeableness, we create two personas, referred to as High Agreeableness and Low Agreeableness. 

\paragraph{Details of Qualitative Shifts. }Each model rates every scenario once in the neutral
condition and once under each of the ten personality settings: the five traits of
the Five-Factor model, each at a high and a low level. We compare every personality setting with the same model's neutral ratings, on the same scenarios and the same 17 appraisal dimensions, at the three levels used for the third-person comparison in Appendix~\ref{sec:app_third_person}. 

First, for each appraisal dimension, with the ratings across all emotions considered together, we test whether the distribution of that dimension's ratings changes, using the Maximum Mean Discrepancy (MMD) metric, with a permutation $p$-value over 500 permutations. Second, for each emotion, we apply the same test to the complete 17-dimension representation, with each dimension standardized across all responses obtained in the two conditions. Third, for each emotion and dimension separately, we use a directional Mann-Whitney test with Cliff's delta, which allows interpreting the direction of the change. In our notation of the delta, a positive value means the neutral response is higher than the personality-conditioned response. Within each level of each comparison, we control the false discovery rate at $5\%$ with the Benjamini-Hochberg procedure~\citep{benjamini1995controlling}. Because the number of ratings is the same in every comparison, we summarize the qualitative shifts by the percentage of tests that are significant, which is comparable across models, traits, emotions, and dimensions. We present our main findings and observations as follows. 

\paragraph{The appraisal distribution from every model changes, and the low level of a trait changes more than the high level.} Personality framing shifts the appraisal-based representation of emotions for all eight models. Averaged over the ten settings, the percentage of emotions whose complete profile changes ranges from $0.59$ for GPT-5.6 Luna to $0.91$ for DeepSeek R1, and the share of individual dimensions that change from $0.40$ to $0.86$. Counting the settings in which more than half of the 17 dimensions change, DeepSeek R1 and QwQ 32B reach 9 of 10, Gemini 2.5 Flash and Muse Glimmer 30B 8, and GPT-5.6 Luna only 2. Conditioning on personality thus leads to a much larger effect than third-person framing or nationality-based personas. The traits differ from one another (Table~\ref{tab:persona_trait}): neuroticism changes the most, at $0.48$ of all emotion-dimension pairs, and extraversion the least, at $0.23$. The difference between the two levels of a trait is larger than the difference between traits: the low level changes $0.45$ of the pairs and the high level $0.21$. Openness is the extreme case, at $0.47$ against $0.04$, so
instructing a model to be open leaves its appraisals almost unchanged (aligned with default model behavior), while instructing it to be closed changes nearly half of them. This is in contrast with results on default personality-based appraisal bias recovered in older model families~\citep{tak2025aware}, where models concurred more with traits of low power, low life satisfaction, low self-efficacy, and high agreeableness. Neuroticism is the only trait whose two levels are alike ($0.49$ and $0.47$).

\paragraph{The emotions that change most are those involving an obstacle, and
valence does not separate them.} Grouping the fifteen emotions by valence does not show any difference at the level of individual emotion-dimension pairs: $0.32$ of pairs change for the nine negative emotions and $0.31$ for the four
positive ones, with the two remaining emotions, challenge and surprise, at $0.43$. When comparing the complete profiles, the negative emotions do change more often ($0.80$ against $0.61$), which means the changes to positive emotions are spread thinly enough across dimensions to be harder to detect as a set. 

The extent to which representations for individual emotions change provides further insights. The emotions that change most are challenge ($0.55$ of pairs), anger ($0.42$) and frustration ($0.41$), each of which could involve an obstacle the person is dealing with. The emotions that change least are disgust ($0.20$), fear ($0.25$) and sadness
($0.28$), which do not directly involve an obstacle in the goal-path. We therefore find no evidence that valence governs how much a personality condition shifts an emotion's appraisal profile.

\paragraph{The change is concentrated in the dimensions describing engagement and effort.} We further examine how individual dimensions differ from each other (Table~\ref{tab:persona_dim}). Four dimensions show large percentages/shares of change: consideration ($0.58$ of pairs), understanding ($0.58$), attention ($0.48$) and exertion ($0.45$). At the other end, self-responsibility ($0.15$), other-control ($0.18$) and fairness ($0.18$) barely move. The two coping-related dimensions, effort and exertion, also change more than the rest, at $0.40$ against $0.32$, with a mean absolute Cliff's delta of $0.42$ against $0.34$. 

The direction of the change depends on the level of the trait more than on the axis of the trait itself. Under the low level of a trait (e.g., low openness, low neuroticism, etc.) the neutral rating is almost always the higher one: averaged over the five traits and the eight models, the Cliff's delta is $+0.50$ for consideration, $+0.40$ for attention, $+0.34$ for exertion, and $+0.28$ for effort, so a model told it is calm, careless, or disagreeable judges the same situation to call for less attention and less effort. Under the high level, the direction reverses ($-0.13$, $-0.01$, $-0.12$ and $-0.08$ respectively for the above-mentioned dimensions). The largest single reversal is expectedly for the trait of neuroticism. Under high neuroticism, every one of the eight models judges situations to require more effort ($-0.42$) and more exertion ($-0.45$) than it does in the neutral condition, and these are the largest shifts of any setting on these two dimensions. High conscientiousness moves effort the same way in all eight models, by less ($-0.09$ for effort and $-0.14$ for exertion), and high agreeableness raises reported understanding ($-0.26$) and self-control ($-0.23$). 

\begin{table}[t]
\centering
\small
\setlength{\tabcolsep}{6pt}
\begin{tabular}{@{}lcccccc@{}}
\toprule
 & \multicolumn{2}{c}{Dimensions} & \multicolumn{2}{c}{Emotions}
 & \multicolumn{2}{c}{Emotion $\times$ dimension} \\
\cmidrule(lr){2-3}\cmidrule(lr){4-5}\cmidrule(lr){6-7}
Trait & High & Low & High & Low & High & Low \\
\midrule
Neuroticism       & 0.77 & 0.88 & 0.94 & 0.93 & 0.47 & 0.49 \\
Conscientiousness & 0.57 & 0.84 & 0.63 & 0.98 & 0.17 & 0.52 \\
Agreeableness     & 0.43 & 0.85 & 0.55 & 0.97 & 0.15 & 0.53 \\
Openness          & 0.11 & 0.83 & 0.16 & 0.96 & 0.04 & 0.47 \\
Extraversion      & 0.49 & 0.59 & 0.56 & 0.78 & 0.20 & 0.26 \\
\bottomrule
\end{tabular}
\caption{Share of tests that show a significant change from the neutral
condition, averaged over the eight models, for each trait at its high and low
level. Dimensions: the 17 per-dimension tests, with emotions pooled. Emotions:
the 15 whole-profile tests. Emotion $\times$ dimension: the 255 directional
tests. All shares are after Benjamini-Hochberg correction within the level of
each comparison.}
\label{tab:persona_trait}
\end{table}

\begin{table}[t]
\centering
\small
\setlength{\tabcolsep}{6pt}
\begin{tabular}{@{}lcc@{\hskip 20pt}lcc@{}}
\toprule
Dimension & Share & $\delta$ & Dimension & Share & $\delta$ \\
\midrule
Consideration        & 0.58 & $+0.19$ & Obstacles            & 0.32 & $+0.02$ \\
Understanding        & 0.58 & $+0.01$ & Enjoyment            & 0.32 & $+0.05$ \\
Attention            & 0.48 & $+0.22$ & Cheated              & 0.24 & $+0.03$ \\
Exertion             & 0.45 & $+0.12$ & Certainty            & 0.22 & $+0.02$ \\
Self-control         & 0.39 & $+0.01$ & Situational control  & 0.21 & $-0.03$ \\
Problem              & 0.38 & $+0.05$ & Other-responsibility & 0.19 & $+0.01$ \\
Pleasantness         & 0.37 & $+0.00$ & Fairness             & 0.18 & $+0.04$ \\
Effort               & 0.36 & $+0.13$ & Other-control        & 0.18 & $+0.01$ \\
                     &      &         & Self-responsibility  & 0.15 & $+0.06$ \\
\bottomrule
\end{tabular}
\caption{Change by appraisal dimension, averaged over the eight models, the ten
personality settings and the fifteen emotions. Share: the fraction of
emotion-and-dimension tests that are significant. $\delta$: the mean Cliff's
delta, written so that a positive value means the neutral rating is higher than
the rating under a persona.}
\label{tab:persona_dim}
\end{table}

\paragraph{Details of shift in \textit{directional} alignment.} Although our central evaluation objective remains \textit{structural} alignment, for the personality-conditioning experiments, we examine \textit{directional} alignment too, to ask whether the slight change toward lesser discreteness with personality affects directional alignment too.

Personality framing indeed moves model responses away from the human responses, and weakens their agreement with appraisal theory. We measure the first as a distance: for one emotion, the multivariate 2-Wasserstein distance between the model's cloud of scenario profiles for that emotion and the human cloud for the same emotion, with all profiles standardized on one common scale, averaged over the fifteen emotions. We measure the second as the correlation between the model's appraisal-emotion associations and the reference values from the meta-analysis from Appendix~\ref{sec:app_agreement_theory}. 

In the neutral condition, the Wasserstein distance of model representations from humans is $3.45$ on average, ranging from $3.10$ for
GPT-5.6 Luna to $4.10$ for Gemini 2.5 Flash, and their agreement with the theoretical predictions is $0.69$. Under a personality condition, the distance to the human ratings
increases in $81\%$ of all possible model-personality condition pairs, by $0.25$ on average (paired Wilcoxon signed-rank test over the eight models, $p=0.016$). The only exception is GPT-5.6 Luna, for which the distance to the human distribution decreases on average by $0.01$. 

Agreement with the theoretical predictions reduces in $65\%$ of the combinations. The two measures order the settings in the same way as the qualitative changes reported above (Table~\ref{tab:persona_alignment}): low agreeableness shifts the models furthest from the human rating distribution ($+0.81$) and reduces agreement with theory the most ($-0.18$), followed by low conscientiousness, high neuroticism and low openness, while low extraversion is the only setting that makes the models marginally closer to humans ($-0.07$). However, these changes are bounded. Considering a single setting at a time, no setting survives correction across the ten (smallest Holm-corrected $p=0.078$), so the evidence of shifting directional alignment holds for personality framing as a whole
rather than for a single given persona.

 
\begin{table}[t]
\centering
\small
\setlength{\tabcolsep}{6pt}
\begin{tabular}{@{}lcccc@{}}
\toprule
 & \multicolumn{2}{c}{Distance to the human ratings}
 & \multicolumn{2}{c}{Agreement with theory} \\
\cmidrule(lr){2-3}\cmidrule(lr){4-5}
Trait & High & Low & High & Low \\
\midrule
Agreeableness     & $+0.05$ & $+0.81$ & $-0.00$ & $-0.18$ \\
Conscientiousness & $+0.07$ & $+0.47$ & $-0.01$ & $-0.08$ \\
Neuroticism       & $+0.44$ & $+0.11$ & $+0.01$ & $-0.04$ \\
Openness          & $+0.09$ & $+0.41$ & $+0.01$ & $-0.06$ \\
Extraversion      & $+0.15$ & $-0.07$ & $-0.01$ & $-0.01$ \\
\bottomrule
\end{tabular}
\caption{Change from the neutral condition under each personality setting,
averaged over the eight models. Distance to the human ratings: the mean $W_2$
distance between the model's and the human profiles of an emotion, averaged over
the fifteen emotions; the neutral value is $3.45$ and a positive change means
the model moved further from the human ratings. Agreement with theory: the
correlation between the model's appraisal-emotion associations and the
meta-analytic reference values; the neutral value is $0.69$ and a negative
change means weaker agreement.}
\label{tab:persona_alignment}
\end{table}

\section{Changes with Post-Training}
\label{sec:app_post_training}

\paragraph{Details of the Setup.} In this section, we provide additional details of the setup we use for investigating a series of Olmo checkpoints~\citep{olmo2025olmo}, across different training stages: 

\paragraph{Checkpoints.} Our analyses uses six different checkpoints of the same 7B model,
from the Olmo 3 family~\citep{olmo2025olmo}, each of which we access directly from Hugging Face, using the following model identifiers: 
\begin{itemize}
    \item Olmo 3 Base Stage 1: a pre-training checkpoint, referred to as \textit{Pretrain} in our analyses, and accessed through \texttt{stage1-step999000} branch of the model \texttt{allenai/Olmo-3-1025-7B}.
    \item Olmo 3 Base Stage 2: a mid-training checkpoint, obtained after applying specific training curriculum using reasoning-heavy data, referred to as \textit{Midtrain}, accessed through \texttt{stage2-step9000} branch of the model \texttt{allenai/Olmo-3-1025-7B}.
    \item Olmo 3 Base: the base model obtained after training on the previous checkpoint to increase the context window, referred to as \textit{Base} in our analyses, accessed through the default main branch for the model \texttt{allenai/Olmo-3-1025-7B}. 
    \item Olmo 3 Instruct SFT: the model obtained after the base model goes through supervised fine-tuning for both long chain-of-thought reasoning and instruction-following and chat behavior, referred to as \textit{SFT} in our analyses. This is accessed using the following model ID: \texttt{allenai/Olmo-3-7B-Instruct-SFT}. 
    \item Olmo 3 Instruct DPO: the model obtained after the instruct SFT model goes through Direct Preference Optimization~\citep{rafailov2023direct}, referred to as \textit{DPO} in our analyses. This is accessed through the following model ID: \texttt{allenai/Olmo-3-7B-Instruct-DPO}. 
    \item Olmo 3 Instruct: the model obtained after the above DPO-tuned model undergoes reinforcement learning with verifiable rewards, specific to math, coding, and instruction-following. This is referred to in our analyses as \texttt{Instruct}, and accessed from: \texttt{allenai/Olmo-3-7B-Instruct}.
\end{itemize}
Every model checkpoint is queried with the same prompt, isolating any observed difference between them to the type of post-training employed.

\paragraph{Exclusion of reasoning checkpoints and larger sizes.} Olmo 3 also releases a corresponding set of checkpoints, where all models are post-trained for enhanced \textit{reasoning}---training models to produce long chains of reasoning before answering. We exclude the reasoning checkpoints for three primary reasons. Firstly, our motivation to examine model checkpoints across training stages stems from existing evidence that alignment post-training adversely impacts \textit{output} diversity~\citep{kirk2024understanding}. Inspired by these findings, we use the alignment post-trained checkpoints to study whether they also induce discreteness and homogeneity at the \textit{concept level}. Second, reasoning models produce reasoning chains \textit{prior} to their final response, making it difficult to distinguish whether any change is due to the type of post-training, or just focused pre-response deliberation. Finally, the length of reasoning generated by these models warrant a higher output token limit, compared to the base or instruction-tuned models, making the comparison unfair. We find this through initial experiments with three reasoning checkpoints from Olmo 3 7B~\citep{olmo2025olmo}: Olmo 3 Think SFT, Olmo 3 Think DPO, and Olmo 3 Think. On the same prompts, the median response is 855, 1131, and 1678 tokens, respectively, for the above checkpoints, against 2 tokens
for the corresponding instruction checkpoint. With a budget of 2048 tokens, $32\%$ of the responses from Olmo 3 Think SFT and $10\%$ of responses from Olmo 3 Think end before the model has finished reasoning, against $2\%$ for the preference-optimized model. Thus, to maintain a controlled, fair comparison, we use only the instruction and preference-optimized ladder of models. For the same reason, we only include the 7B family of models, as the larger model checkpoints (32B) are only released for the reasoning ladder.

\paragraph{Controlling for Format Adherence with Base Models.} Given that this analysis compares responses across base and instruction-tuned models, we use two measures to ensure that base model checkpoints are not penalized unfairly for not producing responses in a specific format. 

Firstly, we adapt our prompt structure slightly to ensure the comparison is fair. A base model has not been trained to follow instructions, and the appraisal prompt used for the other models asks for a JSON object. Scoring such a prompt would confound whether a checkpoint can produce the required format with its appraisal rating. We therefore
use a rating-only version of the same prompt, produced from the original by making three changes: the free-text question about which emotion would be felt is removed, the block specifying the JSON answer format is removed, and the two lines \emph{Respond with only the number, and nothing else.} and \emph{Answer:}
are appended. Within the remaining question, \emph{Answer with a text explanation, and a number} becomes \emph{Answer with a number}. Everything else, including the scenario, the emotion sentence and the numeric scale, is unchanged. 

Secondly, we read each rating from the model's probability distribution over the eleven scale points rather than from the number it produces as its final output. We do this for two main reasons. First, it ensures that a checkpoint is not penalized for a response that does not follow the requested format: a base model that continues the prompt instead of answering still yields a rating, so the comparison across checkpoints measures the appraisal rather than the ability to produce the required format. Second, because the decisiveness of the response is itself the mediator we examine (as shown in the main body), the rating itself must not depend on how decisive the model is. We therefore report the most probable value of the distribution, which is an integer on the same scale as the ratings given by the other models and by the human participants, while ensuring that changes across the checkpoints reflect differences only in post-training strategies and not response format adherence. 
 
\paragraph{Generation settings.} All checkpoints are served with the vLLM library, on a single A100 GPU on a shared Slurm cluster. In the direct answering mode used for all six checkpoints, at most 8 tokens are generated per prompt, which is enough to reach the rating. The base and intermediate checkpoints are prompted as plain text completions, with a single trailing space so that the next token is the digit itself. The instruction-tuned checkpoints are prompted through their chat template, where the candidate forms carry no leading space. 


\section{Prompt Templates}
\label{sec:app_prompt_templates}

\textbf{Prompt Templates. }Next, we provide full descriptions of the exact prompt templates used in our study for evaluation of LLMs. Note that the exact verbatim prompts are provided to the human participants as well, barring information irrelevant for humans, for example, output format requirements for LLMs. 

\begin{promptbox}[Prompt template: Main (self)]
Imagine that you are a person, capable of feeling emotions. Respond to the following prompt as the person. \\
Imagine that you recently <scenario>. And you experienced the following emotion: <emotion>.  Answer the following questions. \\
Question 1: What emotion would you feel in the situation? Answer using a single sentence. \\
Question 2: <appraisal question> \\
Answer in a JSON format as follows: \\
\{ \\
\hspace*{1em}"Question 1": <single sentence response about emotion felt>, \\
\hspace*{1em}"Question 2": [<text response for question 2>, <number response for question 2>] \\
\}
\end{promptbox}

\begin{promptbox}[Prompt template: Third-person]
Consider a person, capable of feeling emotions. Answer the following questions about this person. \\
Imagine that a person recently <scenario>. And they experienced the following emotion: <emotion>.  Answer the following questions. \\
Question 1: What emotion would this person feel in the situation? Answer using a single sentence. \\
Question 2: <appraisal question, phrased in the third person> \\
Answer in a JSON format as follows: \\
\{ \\
\hspace*{1em}"Question 1": <single sentence response about emotion felt>, \\
\hspace*{1em}"Question 2": [<text response for question 2>, <number response for question 2>] \\
\}
\end{promptbox}
 
\begin{promptbox}[Prompt template: Cultural persona]
Imagine that you are a person, capable of feeling emotions. Respond to the following prompt as the following person: You are a person from <cultural information>, living in <cultural information>. Imagine that you recently <scenario>. And you experienced the following emotion: <emotion>.  Answer the following questions. \\
Question 1: What emotion would you feel in the situation? Answer using a single sentence. \\
Question 2: <appraisal question> \\
Answer in a JSON format as follows: \\
\{ \\
\hspace*{1em}"Question 1": <single sentence response about emotion felt>, \\
\hspace*{1em}"Question 2": [<text response for question 2>, <number response for question 2>] \\
\}
\end{promptbox}

\begin{promptbox}[Prompt template: Personality persona]
Imagine that you are a person, capable of feeling emotions. Respond to the following prompt as the following person: <personality information>. Imagine that you recently <scenario>. And you experienced the following emotion: <emotion>.  Answer the following questions. \\
Question 1: What emotion would you feel in the situation? Answer using a single sentence. \\
Question 2: <appraisal question> \\
Answer in a JSON format as follows: \\
\{ \\
\hspace*{1em}"Question 1": <single sentence response about emotion felt>, \\
\hspace*{1em}"Question 2": [<text response for question 2>, <number response for question 2>] \\
\}
\end{promptbox}
 
\begin{promptbox}[Prompt template: Pairwise (explicit discrimination)]
Imagine that you are a person, capable of feeling emotions. \\
How important is each of the following cognitive appraisal dimensions for you in distinguishing <emotion A> from <emotion B>? Rate each dimension on a scale of 1 to 5, where 1 means not at all important and 5 means extremely important. \\
Dimensions: pleasantness, enjoyment, consideration, attention, understanding, situational control, self control, other control, certainty, problem, obstacle, cheated, fairness, self responsibility, other responsibility, exertion, effort. \\
Answer strictly in a JSON format as follows: \\
\{ \\
\hspace*{1em}"<Dimension>": <score from 1 to 5>, \\
\hspace*{1em}... \\
\}
\end{promptbox}

\end{document}